\PassOptionsToPackage{numbers, compress}{natbib}
\documentclass{article}
\usepackage{amsmath}  
\usepackage{hyperref}
\usepackage{float}
\usepackage{amssymb}
\usepackage{xcolor}

\usepackage{subcaption}
\usepackage{listings}
\usepackage{xcolor}

\definecolor{codebg}{RGB}{250,250,250}
\definecolor{keywordcolor}{RGB}{160,32,240}
\definecolor{commentcolor}{RGB}{85,107,47}
\definecolor{stringcolor}{RGB}{0,128,128}
\definecolor{typecolor}{RGB}{255,0,0}

\lstdefinelanguage{PythonCustom}{
    language=Python,
    morekeywords={self},             
    keywordstyle=\color{keywordcolor}\bfseries,
    ndkeywords={def, return, class},
    ndkeywordstyle=\color{keywordcolor}\bfseries,
    identifierstyle=\color{black},
    sensitive=true,
    commentstyle=\color{commentcolor}\itshape,
    stringstyle=\color{stringcolor},
    morecomment=[l][\color{gray}]{\#},
    morestring=[b]',
    morestring=[b]"
}

\lstdefinestyle{custompython}{
    language=PythonCustom,
    backgroundcolor=\color{codebg},
    basicstyle=\ttfamily\small,
    breaklines=true,
    frame=single,
    showstringspaces=false,
    columns=fullflexible,
    captionpos=b
}

\usepackage{neurips_2025}

\usepackage[utf8]{inputenc} 
\usepackage[T1]{fontenc}    
\usepackage{hyperref}       
\usepackage{url}            
\usepackage{booktabs}       
\usepackage{amsfonts}       
\usepackage{nicefrac}       
\usepackage{microtype}      
\usepackage{xcolor}         
\usepackage{multirow}
\usepackage{graphicx}
\usepackage{tabularx}
\usepackage{caption}
\usepackage{wrapfig}
\usepackage{subcaption}

\usepackage{natbib}
\title{
FaCTR: Factorized Channel-Temporal Representation Transformers for Efficient Time Series Forecasting
}

\begin{document}

\maketitle

\begin{abstract}
  While Transformers excel in language and vision—where inputs are semantically rich and exhibit univariate dependency structures—their architectural complexity leads to diminishing returns in time series forecasting. Time series data is characterized by low per-timestep information density and complex dependencies across channels and covariates, requiring conditioning on structured variable interactions. To address this mismatch and overparameterization, we propose FaCTR, a lightweight spatiotemporal Transformer with an explicitly structural design. FaCTR injects dynamic, symmetric cross-channel interactions—modeled via a low-rank Factorization Machine—into temporally contextualized patch embeddings through a learnable gating mechanism. It further encodes static and dynamic covariates for multivariate conditioning. Despite its compact design, FaCTR achieves state-of-the-art performance on eleven public forecasting benchmarks spanning both short-term and long-term horizons, with its largest variant using close to only 400K parameters—on average 50× smaller than competitive spatiotemporal transformer baselines. In addition, its structured design enables interpretability through cross-channel influence scores—an essential requirement for real-world decision-making. Finally, FaCTR supports self-supervised pretraining, positioning it as a compact yet versatile foundation for downstream time series tasks.
\end{abstract}

\section{Introduction}

Time series forecasting is central to industrial decision-making, where errors can cause severe financial and operational losses. In energy markets, a mere 1\% error in electricity demand forecasts can lead to daily losses exceeding \$2 million for mid-sized utilities due to real-time price volatility~\cite{tashman2016}. In healthcare, failing to account for irregularly sampled physiological vitals raises ICU mortality by 12–18\%~\cite{che2016}. Retail and supply chain systems face similar vulnerabilities: studies show that excluding promotional calendars and competitor pricing from demand models inflates forecast errors by 37\%---a gap that no amount of historical sales data alone can close~\cite{makridakis2020}. These cases highlight a core principle: accurate forecasting demands models that move beyond autoregressive dynamics to explicitly condition on exogenous variables such as weather, interventions, or policy shifts—factors that fundamentally shape real-world stochastic processes~\cite{box1976}.

This structural challenge stands in contrast to domains like NLP and vision, where Transformers have thrived. Text tokens inherit semantic density from linguistic abstraction: a single word like \emph{bank} encodes multiple meanings via distributed representations~\cite{mikolov}. Vision operates on spatially coherent structures, where local patches can be meaningfully abstracted via convolutional priors~\cite{dosovitskiy2021}. Crucially, both domains exhibit \textbf{univariate dependency structures} (see Appendix~\ref{appendix:univariate} for a detailed breakdown). In contrast, real-world time series data violate these assumptions. This insight is echoed by empirical findings in \cite{zeng2023}, which demonstrates that simple linear models outperform state-of-the-art Transformers on long-horizon benchmarks when covariates are omitted. Similarly, studies such as \cite{ilbert2024, li2024strep, das2023, nie2023patchtst} show that incorporating auxiliary signals---including calendar features, external drivers, and event indicators---consistently improves accuracy, especially under distribution shift.

While modern Transformer architectures have advanced spatial modeling in time series, they often conflate temporal and cross-channel dependencies into a single attention map—a design seen in Spacetimeformer~\cite{wu2021spacetimeformer}. Though expressive, this monolithic structure introduces two inefficiencies. First, it causes a combinatorial explosion in attention space, scaling as \(O(C^2T^2)\) for \(C\) channels and \(T\) timesteps. Second, it overlooks that temporal and cross-channel dependencies arise from distinct generative processes and operate on different timescales. Temporal patterns typically reflect autoregressive dynamics, while cross-channel relationships encode stable structural couplings. Two inductive biases clarify this: \textbf{timescale separation}, where inter-channel dependencies decay faster than temporal ones, and \textbf{structural invariance}, where inter-variable relationships—such as physical laws—remain stable over time (refer Appendix~\ref{appendix:inductivebias} for details). Conflating these leads to representational overreach, reducing statistical efficiency and misallocating model capacity. The challenge isn’t modeling dependencies—it’s modeling the \textit{right} dependencies efficiently.

\begin{table}[h]
\scriptsize
\centering
\caption{Model size comparison (parameter count) across datasets and forecast horizons. All models use a lookback window of 512. Values are reported in number of trainable parameters; \textcolor{blue}{blue} indicates the most compact model per row. FaCTR consistently exhibits the lowest parameter footprint across datasets and horizons. Refer to Appendix~\ref{appendix:paramsapp} for detailed benchmarking with more horizons.}
\label{tab:params}
\begin{tabular}{llrrrrrrr}
\toprule
\textbf{Dataset} & \textbf{Horizon}  & \textbf{FaCTR} & \textbf{PatchTST} & \textbf{DLinear} & \textbf{TSMixer} & \textbf{CrossFormer} & \textbf{iTransformer} & \textbf{ModernTCN} \\
\midrule
\multirow{2}{*}{ETTh2} 
& 96 & \textcolor{blue}{71{,}296} & 115{,}872 & 98{,}496 & 576{,}604 & 11{,}811{,}096 & 224{,}224 & 656{,}260 \\
& 720 & 391{,}408  & 755{,}472 & 738{,}720 & 896{,}620 & 11{,}601{,}248 & \textcolor{blue}{304{,}720} & 4{,}011{,}508 \\
\midrule
\multirow{2}{*}{Weather} 
& 96 & \textcolor{blue}{71{,}744} & 1{,}194{,}336 & 98{,}496 & 1{,}105{,}598 & 11{,}518{,}768 & 4{,}833{,}888 & 2{,}489{,}476 \\
& 720 & \textcolor{blue}{391{,}856} & 6{,}306{,}768 & 738{,}720 & 1{,}425{,}614 & 11{,}667{,}808 & 5{,}154{,}000 & 10{,}268{,}404 \\
\midrule
\multirow{2}{*}{Traffic} 
& 96 & \textcolor{blue}{98{,}656} & 921{,}187 & 98{,}496 & 3{,}042{,}412 & 3{,}589{,}744 & 6{,}411{,}872 & 822{,}756{,}868 \\
& 720 & \textcolor{blue}{418{,}768} & 4{,}276{,}435 & 738{,}720 & 3{,}362{,}428 & 3{,}681{,}440  & 6{,}534{,}992 & 832{,}342{,}132 \\
\midrule
\multirow{2}{*}{Electricity} 
& 96 & \textcolor{blue}{81{,}344} & 1{,}194{,}336 & 98{,}496 & 1{,}266{,}502 & 2{,}938{,}200 & 4{,}833{,}888 & 129{,}146{,}500 \\
& 720 & \textcolor{blue}{401{,}456} & 6{,}306{,}768 & 738{,}720 & 1{,}586{,}518 & 1{,}880{,}992 & 5{,}154{,}000 & 136{,}335{,}604 \\
\bottomrule
\end{tabular}
\end{table}

Another core limitation is \textbf{parameter inefficiency}: many recent architectures allocate disproportionately large capacities relative to forecasting complexity. This manifests in two recurring patterns (see Table~\ref{tab:params}): (\textbf{1}) \textbf{Overparameterization}—ModernTCN~\cite{zhou2024moderntcn} exceeds 136M parameters on Electricity, over 300$\times$ larger than FaCTR; (\textbf{2}) \textbf{Redundancy}—TSMixer~\cite{wu2023tsmixer} uses over 3M parameters on Traffic for marginal gains over 90K-param baselines. These inefficiencies arise from architectural choices such as channel–time separable but dimensionally sensitive mixers in TSMixer, and stacked multi-head attention in PatchTST. Informer~\cite{zhou2021informer} and FEDformer~\cite{zhou2022fedformer}, while introducing sparsity or frequency priors, still rely on large feedforward blocks or global mixing. Spatiotemporal models like CrossFormer~\cite{wu2023crossformer} and STAEFormer~\cite{liu2023staeformer} often exceed millions of parameters, while MTGNN~\cite{xie2020mtgnn}\footnote{Recent models such as \textsc{FourierGNN} have shown improved performance over MTGNN, but are excluded here as they do not evaluate on our benchmark datasets.} and CrossGNN~\cite{gong2021crossgnn} are relatively compact ($\sim$400K parameters), yet lack covariate integration and perform worse in comparison (see Table~\ref{tab:performance}).

\paragraph{Our Approach:}
To address these limitations, we introduce \textbf{FaCTR} (\textit{Factorized Channel-Temporal Representation Transformer})—a lightweight, structurally grounded model that aligns with the generative structure of real-world multivariate forecasting tasks. Despite its compact architecture, FaCTR achieves state-of-the-art forecasting accuracy across both short- and long-term horizons on eleven standard benchmarks (see Table~\ref{tab:performance} and Table~\ref{tab:performance_short}). FaCTR is built on four architectural principles: 

\vspace{1mm}
\begin{itemize}
    \item \textbf{Structural Disentanglement with Full ERF Coverage:} 
    FaCTR decouples forecasting into three structurally distinct pathways: temporal self-attention per variable stream, cross-channel interaction via a low-rank Factorization Machine (FM), and embedding-wise mixing through a lightweight MLP. The temporal module operates patchwise and channel-independently, inheriting PatchTST’s Effective Receptive Field (ERF) benefits—$\sim$98\% coverage, uniform signal propagation, and stable long-horizon gradients—unlike sparse variants~\cite{zhou2021informer, zhou2022fedformer, jin2020} with 65–73\% coverage~\cite{nie2023patchtst}. Cross-channel ERF is complete by design, with FM capturing all pairwise variable interactions at each patch. The MLP similarly spans the full embedding for each channel, ensuring full latent coverage.

    \item \textbf{Principled Cross-Channel Modeling:} At the core of FaCTR is a novel application of the \textit{Factorization Machine} to time series forecasting. Originally developed for high-sparsity recommendation tasks~\cite{rendle2010fm}, FM models pairwise feature interactions via symmetric low-rank decomposition. We repurpose this mechanism to capture variable-wise dependencies across channels—treating inter-variable structure as an implicit interaction graph learned from per-patch channel embeddings enriched with static and dynamic covariates. This design replaces full-rank spatial attention with inductive priors aligned with sparsity, symmetry, and time-invariant structure. FM’s scalability makes it well-suited to high-cardinality forecasting regimes—such as item–customer–region combinations in retail, where entities number in the hundreds of thousands and exhibit sparse cross-dependencies. To our knowledge, FaCTR is the first architecture to integrate FM into multivariate forecasting, leveraging its natural compatibility with structured cross-channel interactions.

    \item \textbf{Compact and Scalable Design:} 
    FaCTR achieves up to $\sim$50$\times$ parameter savings on average compared to spatiotemporal Transformer baselines (see Appendix~\ref{appendix:efficiencymul}), while maintaining high representational capacity. By decoupling temporal and cross-channel modeling, it avoids the $\mathcal{O}((C \cdot L)^2)$ cost of dense spatiotemporal attention. This leads to a total sub-quadratic complexity of $\mathcal{O}(C N (C + N))$ (see Appendix~\ref{appendix:time-complexity} for a full derivation). Here, $L$ is the original sequence length and $N$ is the number of non-overlapping patches. A comparative analysis of training time and peak memory usage is provided in Appendix~\ref{appendix:efficiency}.

    \item \textbf{Interpretability through Structured Attention:} FaCTR exposes interpretable representations at two levels: \textit{temporal attention maps} provide autoregressive attribution per channel across patches, while \textit{FM scores} reveal dynamic pairwise channel interactions at each patch. 

\end{itemize}
\section{Related Work}

\paragraph{Time Series Forecasting.}
Modern forecasting has evolved from classical statistical models such as ARIMA~\cite{box1976} to deep learning architectures like RNNs~\cite{che2016} and TCNs~\cite{zhou2024moderntcn}, which introduced greater modeling flexibility but struggled with long-term dependencies. Transformers marked a major shift: Informer~\cite{zhou2021informer} introduced probabilistic sparsity for scalable attention, Autoformer~\cite{wu2021autoformer} and FEDformer~\cite{zhou2022fedformer} employed seasonal-trend decomposition, and Pyraformer~\cite{liu2022pyraformer} leveraged a pyramidal hierarchy. However, these models lack explicit channel mixing, are parameter-inefficient, and among those that aim to reduce the quadratic complexity of attention, many offer limited effective receptive field (ERF) coverage. PatchTST~\cite{nie2023patchtst} improved ERF and temporal fidelity via patchwise modeling but entirely ignored cross-channel structure. DeformTime~\cite{shu2025deformtime} introduced deformable attention to handle variable dependencies, at the cost of increased architectural complexity. More compact MLP-style models such as TSMixer~\cite{wu2023tsmixer} and LightTS~\cite{zhang2023lightts} improve parameter efficiency but lack a principled mechanism for inter-channel interaction beyond generic mixing through feedforward layers. Graph-based approaches like CrossGNN~\cite{gong2021crossgnn} and MTGNN~\cite{xie2020mtgnn} model inter-channel dependencies more explicitly but typically underperform Transformer-based models on standard long-term forecasting benchmarks. SAMFormer~\cite{ilbert2024}, while parameter-efficient, does not perform explicit temporal mixing via attention. Crucially, none of the above offer a scalable and interpretable mechanism to model sparse but important inter-channel dependencies in a structured and computationally efficient way.

\paragraph{Factorization Machines for Structured Interaction Modeling.}
Factorization Machines (FM)~\cite{rendle2010fm} offer an efficient inductive bias for modeling pairwise feature interactions under sparsity. While widely adopted in recommendation systems for learning implicit structure across high-cardinality variables, their utility in time series remains largely untapped. Neural variants like DeepFM~\cite{guo2017deepfm} and xDeepFM~\cite{lian2018xdeepfm} demonstrate that FM layers can be seamlessly embedded within deep models to learn rich, structured dependencies. Related hybrids such as Attentional FM~\cite{xiao2017afm} and Deep\&Cross~\cite{wang2017dcn} bridge attention and explicit cross modeling but have not been applied in temporal settings. FaCTR integrates a low-rank FM layer into a patch-based temporal encoder to explicitly model inter-channel structure—a sparse, often stable dependency type common in real-world time series. This disentangled design enables FaCTR to avoid the drawbacks of dense spatial attention while preserving interpretability and scaling to high-dimensional settings.

\section{Model Architecture}
\label{proposed_approach}

\begin{figure}[htbp]
  \centering
  \includegraphics[width=1.0\linewidth]{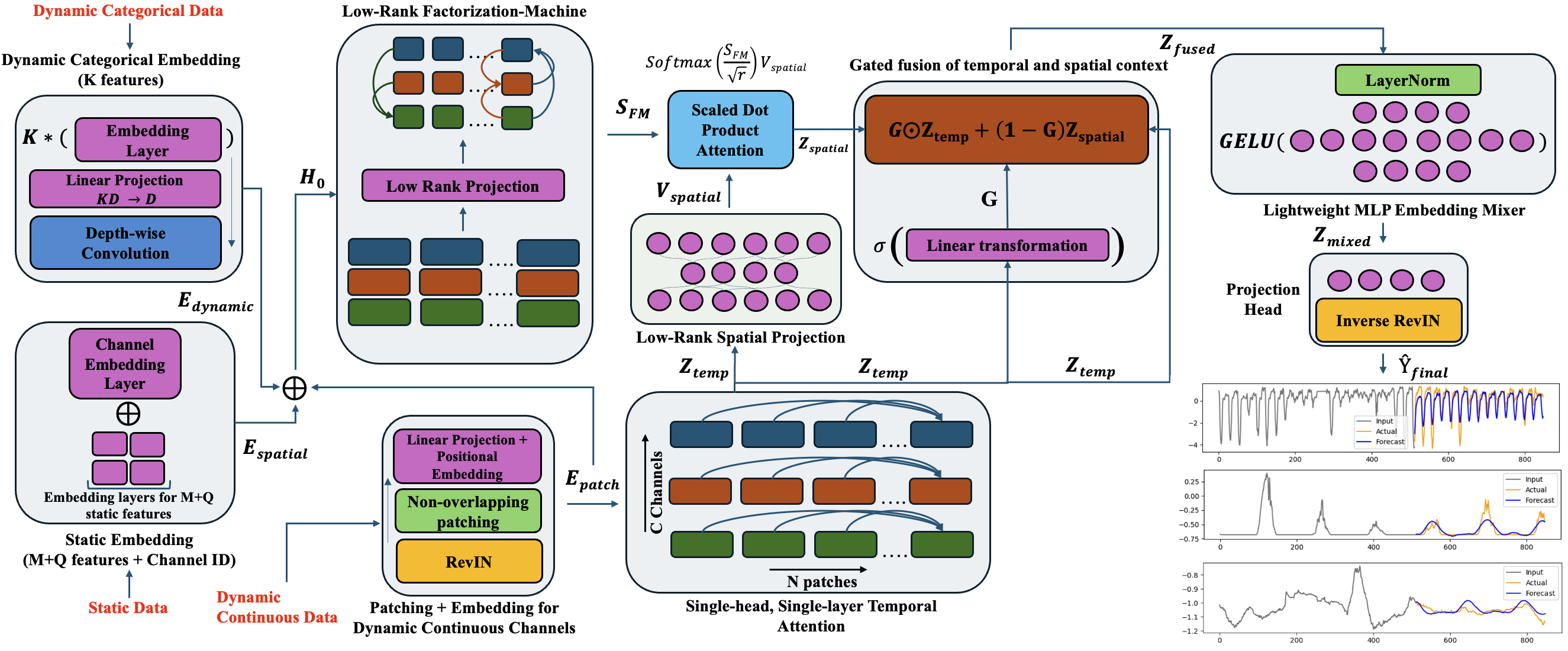}
  \caption{FaCTR architecture: A disentangled spatiotemporal model where temporal attention, low-rank Factorization Machine-based channel mixing, and embedding-wise MLP operate independently along the patch, channel, and feature axes. This structure enables scalable, interpretable forecasting.}

  \label{fig:your_label}
\end{figure}

\paragraph{Problem Formulation}
Let the input multivariate time series be denoted by $\mathbf{X} \in \mathbb{R}^{B \times C \times L}$, where $B$ is the batch size, $C$ is the number of channels, and $L$ is the input sequence length. The forecasting task is to predict the next $T$ steps, yielding $\hat{\mathbf{Y}} \in \mathbb{R}^{B \times C \times T}$. P refers to patch length, S refers to stride and N refers to the number of patches. D is the hidden dimension of the model.

Figure~\ref{fig:your_label} provides an overview of FaCTR. The architecture is composed of the following: \textit{Patching and Covariate Alignment}; \textit{Structured Spatiotemporal Attention}, which includes a \textbf{single-head} temporal self-attention mechanism, a low-rank FM-based cross-channel interaction module, and a learnable gating unit for fusion; \textit{Embedding Mixing}; and a \textit{Projection Head} for decoding to the forecast horizon. For more details on the architecture and design, please refer to Appendix \ref{appendix:designrat}.

\subsection{Patching and Covariate Alignment}

\paragraph{Normalization.}
Each dynamic continuous-valued time series channel is first normalized independently using RevIN~\cite{kim2022revin} (see Appendix~\ref{appendix:revin} for details) to mitigate non-stationarity and distributional drift across training and inference. For instance, in retail, while the focus is on item-level demand, auxiliary series like pricing can be treated as channels and normalized similarly. Static continuous features are z-score normalized per channel using training-set statistics.

\paragraph{Patching.} We segment $\mathbf{X} \in \mathbb{R}^{B \times C \times L}$ into $N = \left\lfloor \frac{L - P}{S} \right\rfloor + 1$ non-overlapping patches of shape $\mathbf{X}_{\text{patch}} \in \mathbb{R}^{B \times C \times N \times P}$, and apply a shared linear projection $\mathbb{R}^{P} \rightarrow \mathbb{R}^{D}$ followed by patch-wise positional encoding to obtain $\mathbf{E}_{\text{patch}} = \text{Linear}(\mathbf{X}_{\text{patch}}) + \mathbf{E}_{\text{pos}}$, where $\mathbf{E}_{\text{pos}} \in \mathbb{R}^{1 \times 1 \times N \times D}$. While our formulation supports overlapping or non-overlapping patches, we use $S = P$ for efficiency.

We encode all static and dynamic covariates—both categorical and continuous—using dedicated embedding layers, producing two auxiliary representations: \textbf{static embeddings}, which capture per-channel identity and metadata (e.g., category, location, shelf life), and \textbf{dynamic categorical embeddings}, which encode time-varying exogenous signals such as calendar features. Each embedding is temporally aligned with the input patches and broadcast across the spatial dimension when needed, providing structured inductive priors for cross-channel modeling by the FM and eliminating the need for manual feature engineering or handcrafted interactions. 

\paragraph{Static Embeddings.}
Each channel $c \in \{1, \dots, C\}$ is assigned a unique identifier and mapped via a learnable embedding $\text{Embed}_{\text{channel}}(c) = \mathbf{e}_c \in \mathbb{R}^D$. Suppose each channel also has $M$ categorical static attributes (e.g., product type, location), with each attribute $z_c^{(m)} \in \mathbb{Z}$ embedded via $\text{Embed}_m : \mathbb{Z} \rightarrow \mathbb{R}^D$. The static categorical embedding for channel $c$ is then $\mathbf{e}_c^{\text{static-cat}} = \sum_{m=1}^M \text{Embed}_m(z_c^{(m)})$. In addition, we allow for $Q$ static normalized continuous features per channel (e.g., average price, shelf life) $\mathbf{x}_c^{\text{cont}} \in \mathbb{R}^Q$ which are projected using a shared linear layer $\mathbf{W}_{\text{cont}} \in \mathbb{R}^{Q \times D}$, yielding continuous embeddings $\mathbf{e}_c^{\text{static-cont}} = \mathbf{W}_{\text{cont}} \cdot \mathbf{x}_c^{\text{cont}}$. Each of the above components lies in $\mathbb{R}^D$, and the full static embedding for channel $c$ is obtained by summing them: \(
\mathbf{e}_c^{\text{static}} = \mathbf{e}_c + \mathbf{e}_c^{\text{static-cat}} + \mathbf{e}_c^{\text{static-cont}}.
\) Stacking across all channels gives $\mathbf{E}_{\text{spatial}} \in \mathbb{R}^{C \times D}$, which is expanded across the batch and patch dimensions to yield $\mathbf{E}_{\text{spatial}}^{\text{expanded}} \in \mathbb{R}^{B \times C \times N \times D}$.

\paragraph{Dynamic Categorical Embeddings.}
Let $\mathbf{F}_{\text{dyn}} \in \mathbb{Z}^{B \times C \times L \times K}$ denote $K$ time-varying categorical covariates per channel, e.g., calendar features like day-of-week. For each $k \in [1, K]$, the feature $\mathbf{F}_{\text{dyn}}^{(k)} \in \mathbb{Z}^{B \times C \times L}$ is mapped via a learnable embedding $\text{Embed}_k : \mathbb{Z} \rightarrow \mathbb{R}^D$. After concatenation across features, we obtain the raw embedding tensor $\mathbf{E}_{\text{dyn}}^{\text{raw}} \in \mathbb{R}^{B \times C \times L \times (K \cdot D)}$, which is then projected to $\mathbb{R}^{B \times C \times L \times D}$ via a linear layer. This sequence is passed through a depthwise 1D convolution in the temporal direction with kernel size and stride equal to patch length $P$ and stride $S$, producing $\mathbf{E}_{\text{dyn}} \in \mathbb{R}^{B \times C \times N \times D}$.

\paragraph{Context-Aware Prior Representation.}
The covariate-informed prior is computed as the elementwise sum $\mathbf{Z}_{\text{context}} = \mathbf{E}_{\text{spatial}}^{\text{expanded}} + \mathbf{E}_{\text{dyn}} \in \mathbb{R}^{B \times C \times N \times D}$ and is added to the patch embeddings of the continuous variables, yielding $\mathbf{H}_0 = \mathbf{E}_{\text{patch}} + \mathbf{Z}_{\text{context}} \in \mathbb{R}^{B \times C \times N \times D}$. This fused representation $\mathbf{H}_0$—now temporally localized and covariate-informed—is fed exclusively into the \textbf{Factorization Machine} for cross-channel interaction modeling.

\subsection{Structured Spatiotemporal Attention}

Following the covariate-informed representation $\mathbf{H}_0 \in \mathbb{R}^{B \times C \times N \times D}$ from the prior stage, FaCTR models dependencies across both the temporal and channel dimensions with a disentangled architecture. 

\paragraph{Temporal Attention.}
We begin by modeling sequential dependencies within each channel independently using a single-head scaled dot-product attention mechanism~\cite{vaswani2017}. Let $\mathbf{E}_{\text{patch}} \in \mathbb{R}^{B \times C \times N \times D}$ denote the input patch embeddings. Temporal attention is applied along the $N$ patch positions for each channel: $\text{Attention}(\mathbf{Q}, \mathbf{K}, \mathbf{V}) = \text{Softmax}\left( \frac{\mathbf{Q} \mathbf{K}^\top}{\sqrt{D}} \right)\mathbf{V}$, where queries, keys, and values are learned projections of $\mathbf{E}_{\text{patch}}$. The output is a temporally contextualized representation: $\mathbf{Z}_{\text{temp}} = \text{TemporalAttn}(\mathbf{E}_{\text{patch}}) \in \mathbb{R}^{B \times C \times N \times D}$. 

\paragraph{Low-Rank Factorized Channel Interaction.}
Multivariate time series exhibit structured but sparse statistical dependencies between channels (e.g., retail items). Empirically, the mutual information between most variable pairs is negligible—a phenomenon detailed in Appendix~\ref{appendix:mi-sparsity}. This motivates modeling inter-channel structure with inductive biases toward sparsity and low-rank geometry. Rather than relying on costly pairwise token interactions as in traditional self-attention, FaCTR employs a low-rank Factorization Machine (FM) to capture cross-channel dependencies efficiently.

From the fused representation $\mathbf{H}_0 \in \mathbb{R}^{B \times C \times N \times D}$ defined earlier, we compute a per-patch, per-batch channel similarity matrix via inner product: letting $\mathbf{h}_{i,n} \in \mathbb{R}^{D}$ denote the embedding of channel $i$ at patch position $n$, we define $\text{FM}_{i,j}^{(b,n)} = \langle \mathbf{h}_{i,n}, \mathbf{h}_{j,n} \rangle$, yielding a similarity tensor $\mathbf{S}_{\text{FM}} \in \mathbb{R}^{B \times C \times C \times N}$. Rather than computing a full $C \times C$ pairwise similarity matrix via dense channel attention, we approximate it with a rank-$r$ decomposition $\mathbf{S}_{\mathrm{FM}} = \mathbf{V}\mathbf{V}^\top$, where $\mathbf{V} \in \mathbb{R}^{C \times r}$, introducing only $\mathcal{O}(C r)$ learned parameters and imposing a low-rank inductive bias aligned with sparse cross-channel dependencies. This corresponds to the classical FM structure. By the Eckart–Young–Mirsky theorem, this decomposition yields the optimal rank-$r$ approximation under the Frobenius norm: $\min_{\tilde{\mathbf{S}},\, \text{rank}(\tilde{\mathbf{S}})\leq r} \|\mathbf{S} - \tilde{\mathbf{S}}\|_F^2 = \sum_{k=r+1}^{C} \sigma_k^2$. The resulting low-rank inductive bias yields compact—and often time-invariant—representations of variable interactions, aligning well with the sparsity patterns observed in climate, energy, and retail datasets.

To normalize the similarity matrix into a cross-channel influence distribution, we apply softmax over the source (second) channel: $\mathbf{A}_{\text{spatial}} = \text{Softmax}\left(\frac{\mathbf{S}_{\text{FM}}}{\sqrt{r}}\right)$, producing $\mathbf{A}_{\text{spatial}} \in \mathbb{R}^{B \times C \times C \times N}$, which encodes the relative influence of each source channel on every target channel across patches.

\paragraph{Low-Rank Spatial Projection.} 
To apply the FM-based attention weights, we project the temporal embeddings into a subspace compatible with the learned inter-channel structure. The raw temporal outputs—optimized for sequential patterns—are not naturally aligned with the FM geometry. We bridge this mismatch using a two-layer low-rank MLP, applied independently to each channel-patch embedding: $\mathbf{V}_{\text{low}} = \mathbf{W}_{\text{low}} \mathbf{Z}_{\text{temp}}, \quad \mathbf{W}_{\text{low}} \in \mathbb{R}^{D \times r}$, followed by $\mathbf{V}_{\text{spatial}} = \mathbf{W}_{\text{high}} \mathbf{V}_{\text{low}}, \quad \mathbf{W}_{\text{high}} \in \mathbb{R}^{r \times D}$, yielding $\mathbf{V}_{\text{spatial}} \in \mathbb{R}^{B \times C \times N \times D}$. This bottleneck reduces parameter count while aligning the representational spaces of temporal and spatial components, retaining expressivity. As with FM, the rank constraint provides an optimal low-rank approximation under the Frobenius norm.

\paragraph{Gated Fusion of Temporal and Spatial Signals.}
The FM-derived attention weights are used to aggregate information across correlated channels: $\mathbf{Z}_{\text{spatial}} = \mathbf{A}_{\text{spatial}} \cdot \mathbf{V}_{\text{spatial}}, \quad \mathbf{Z}_{\text{spatial}} \in \mathbb{R}^{B \times C \times N \times D}$, where the product is a weighted sum over the source channel dimension. To integrate these globally-aware spatial representations with the localized temporal outputs $\mathbf{Z}_{\text{temp}}$, we introduce a learnable gating mechanism: $\mathbf{G} = \sigma(\mathbf{W}_g \mathbf{Z}_{\text{temp}}), \quad \mathbf{W}_g \in \mathbb{R}^{D \times D}, \quad \mathbf{G} \in \mathbb{R}^{B \times C \times N \times D}$. The fused representation is computed as $\mathbf{Z}_{\text{fused}} = \mathbf{G} \odot \mathbf{Z}_{\text{temp}} + (1 - \mathbf{G}) \odot \mathbf{Z}_{\text{spatial}}$. Since the gating weights depend only on temporally contextualized features, the fusion mechanism is inherently biased toward recency—allowing temporal dynamics to dominate while modulating them with cross-channel signals when informative. This results in a convex, per-location blend of local and global context that is flexible, interpretable, and fully learnable. The disentangled design enables FaCTR to approximate smooth dependency surfaces that flexibly combine autoregressive and spatial influences.

\subsection{Embedding-wise Mixing and Projection Head}

\paragraph{Embedding-wise Mixing.}
The spatiotemporally fused tensor \( \mathbf{Z}_{\text{fused}} \in \mathbb{R}^{B \times C \times N \times D} \) is refined through a lightweight two-layer feedforward network applied at each \( (b, c, n) \) location. Letting \( \mathbf{z}_{bcn} \in \mathbb{R}^D \) denote the embedding at that index, we apply LayerNorm (following the Pre-LN convention), followed by:
\(
\mathbf{h}_{bcn} = \mathbf{z}_{bcn} + \text{MLP}(\mathbf{z}_{bcn}), \quad \text{where} \quad \text{MLP}(\mathbf{z}) = \mathbf{W}_2\, \text{GELU}(\mathbf{W}_1 \mathbf{z} + \mathbf{b}_1) + \mathbf{b}_2
\)
with \( \mathbf{W}_1 \in \mathbb{R}^{D \times 4D} \), \( \mathbf{W}_2 \in \mathbb{R}^{4D \times D} \). This residual block improves gradient flow and expressivity with minimal parameter overhead; dropout is applied between layers during training. We denote the output tensor after embedding-wise mixing as \( \mathbf{Z}_{\text{mixed}} \in \mathbb{R}^{B \times C \times N \times D} \).

\paragraph{Flattening and Forecast Projection.}
The mixed tensor \( \mathbf{Z}_{\text{mixed}} \) is flattened across patches and embedding dimensions as \( \mathbf{Z}_{\text{flat}} = \text{reshape}(\mathbf{Z}_{\text{mixed}}) \in \mathbb{R}^{B \times C \times (N \cdot D)} \), then projected via a channel-specific linear layer:
\(
\hat{\mathbf{Y}} = \mathbf{W}_{\text{proj}} \mathbf{Z}_{\text{flat}} + \mathbf{b}_{\text{proj}} \in \mathbb{R}^{B \times C \times T}
\)
where \( \mathbf{W}_{\text{proj}} \in \mathbb{R}^{(N \cdot D) \times T} \). The output is finally denormalized using the inverse RevIN transform: \( \hat{\mathbf{Y}}_{\text{final}} = \text{RevIN}^{-1}(\hat{\mathbf{Y}}) \).

\subsection{Computational Complexity Summary}
\label{subsec:complexity}
FaCTR has total time complexity  
\(
\mathcal{O}(C N^2 + C^2 N + N T),
\)
aggregating costs from temporal attention ($\mathcal{O}(C N^2)$), FM-based channel mixing ($\mathcal{O}(C^2 N)$), and projection head ($\mathcal{O}(N T)$). A detailed derivation and component-wise breakdown are provided in Appendix~\ref{appendix:time-complexity}.

\section{Experiments}

\subsection{Setup}

\paragraph{Datasets and Evaluation}

We evaluate FaCTR on both short-term and long-term multivariate time series forecasting tasks. For long-term forecasting, we use standard benchmarks across diverse domains—ETT (ETTh1, ETTh2, ETTm1, ETTm2), Weather, Electricity, and Traffic. For short-term forecasting, we adopt the PEMS datasets, widely used in prior studies\footnote{https://drive.google.com/file/d/1l51QsKvQPcqILT3DwfjCgx8Dsg2rpjot/view}. See Appendix~\ref{appendix:datasets} for detailed dataset descriptions. Performance is measured using two standard regression metrics—Mean Squared Error (MSE) and Mean Absolute Error (MAE)—across prediction horizons of 96, 192, 336, and 720 steps for long-term forecasting, and 12, 24, 48, and 96 steps for short-term forecasting, in line with existing literature.

\paragraph{Model Baselines} To evaluate FaCTR, we benchmark against a diverse set of state-of-the-art models spanning major forecasting paradigms. For long-term multivariate time series forecasting, we consider \textbf{Transformer-based} models such as \textsc{FEDformer}~\cite{zhou2022fedformer} (temporal attention) and \textsc{CSFormer}~\cite{zhang2025csformer} (temporal + channel attention), and \textsc{SAMFormer}~\cite{ilbert2024} (channel attention); \textbf{Convolutional} architectures like \textsc{ModernTCN}~\cite{zhou2024moderntcn}; \textbf{Linear} baselines including \textsc{DLinear}~\cite{zeng2023}; \textbf{Patching-based} models like \textsc{PatchTST}~\cite{nie2023patchtst}; \textbf{Mixer-based} architectures such as \textsc{TSMixer}~\cite{wu2023tsmixer}; \textbf{Graph-based} approaches such as \textsc{CrossGNN}~\cite{gong2021crossgnn} and the foundation model \textsc{Moment}~\cite{goswami2024moment}. Both \textsc{SAMFormer} and \textsc{TSMixer} are optimized by \textbf{Structured Aware Minimization (SAM)}~\cite{ilbert2024}. For short-term forecasting, in addition to some of the above, we include additional baselines known to perform well in high-resolution, low-horizon settings such as \textsc{iTransformer}~\cite{liu2024itransformer}, \textsc{SCINet}~\cite{liu2022scinet}, \textsc{CrossFormer}~\cite{wu2023crossformer} respectively. See Appendix~\ref{appendix:modelperf} for more details on baselines.

\paragraph{Implementation Details}
FaCTR is trained using the \textbf{Adam optimizer} with SAM optimization (refer to Appendix~\ref{appendix:sam}), and the \textbf{L2 loss} (mean squared error) as the training objective. To mitigate overfitting, we apply \textbf{early stopping} with a patience of 10 epochs based on validation loss. All hyperparameters were extensively tuned on the validation set (see Appendix~\ref{appendix:hyparams} for details).

\subsection{Results – Multivariate Time Series Forecasting}

\textbf{FaCTR consistently achieves state-of-the-art performance across both short- and long-term forecasting tasks.} As shown in Table~\ref{tab:performance}, in the long-term forecasting setting, it ranks first (lowest error) in 11 out of 28 MSE scores (and 14 for MAE), and second in 13 more for MSE (and 8 for MAE)—demonstrating strong generalization across domains with varying temporal characteristics, from high-frequency signals (\textbf{ETTm2}) to low-frequency ones (\textbf{Weather}). For short-term forecasting, shown in Table~\ref{tab:performance_short}, FaCTR ranks first in 11 out of 16 horizons for MSE (and 11 for MAE), and second in the remaining 5 for both metrics across the PEMS datasets, consistently outperforming or matching leading baselines. It achieves this with significantly fewer parameters than typical baselines—validating its design philosophy of leveraging structural priors over brute-force scale. Furthermore, FaCTR delivers these results with minimal tuning: its architecture remains fixed across all datasets and horizons, underscoring its ease of use and adaptability. All reported results are averaged over three runs to account for variability. The main text highlights the most competitive baselines, with extended comparisons provided in Appendix~\ref{appendix:modelperformance}, where FaCTR consistently emerges as the top performer. Training time and memory consumption are detailed in Appendix~\ref{appendix:efficiency}. Some examples of forecasts generated by the model across different datasets and horizons can be found in ~\ref{appendix:forecasts}.

\begin{table}[ht]
\centering
\caption{Long-term Forecasting performance (MSE and MAE) across models and datasets. Blue highlights denote the best score per row, and red indicates the second-best. FaCTR consistently outperforms all baselines in the majority of settings across both metrics.}
\noindent\resizebox{\textwidth}{!}{%
\label{tab:performance}
\begin{tabular}{llcccccccccccccccccccc}
\toprule
\multirow{2}{*}{\textbf{}} & \multirow{2}{*}{\textbf{Horizon}} & \multicolumn{2}{c}{\textbf{FaCTR}} & \multicolumn{2}{c}{\textbf{PatchTST}} & \multicolumn{2}{c}{\textbf{ModernTCN}} & \multicolumn{2}{c}{\textbf{Dlinear}} & \multicolumn{2}{c}{\textbf{TSMixer}} & \multicolumn{2}{c}{\textbf{SAMFormer}} & \multicolumn{2}{c}{\textbf{CrossGNN}} & \multicolumn{2}{c}{\textbf{CSFormer}} & \multicolumn{2}{c}{\textbf{FedFormer}} & \multicolumn{2}{c}{\textbf{Moment}} \\
\cmidrule(lr){3-4} \cmidrule(lr){5-6} \cmidrule(lr){7-8} \cmidrule(lr){9-10} \cmidrule(lr){11-12} \cmidrule(lr){13-14} \cmidrule(lr){15-16} \cmidrule(lr){17-18} \cmidrule(lr){19-20} \cmidrule(lr){21-22} 
& & MSE & MAE & MSE & MAE & MSE & MAE & MSE & MAE & MSE & MAE & MSE & MAE & MSE & MAE & MSE & MAE & MSE & MAE & MSE & MAE \\ \midrule
Weather & 96 & \textcolor{blue}{0.148} & \textcolor{red}{0.200} & \textcolor{blue}{0.148} & \textcolor{blue}{0.197} & \textcolor{red}{0.153} & 0.210 & 0.171 & 0.230 & 0.189 & 0.242 & 0.197 & 0.249 & 0.159 & 0.218 & 0.168 & 0.217 & 0.238 & 0.314 & 0.154 & 0.209 \\
 & 192 & \textcolor{blue}{0.191} & \textcolor{blue}{0.240} & \textcolor{red}{0.192} & \textcolor{blue}{0.240} & 0.208 & \textcolor{red}{0.256} & 0.216 & 0.274 & 0.228 & 0.272 & 0.235 & 0.277 & 0.211 & 0.266 & 0.213 & \textcolor{red}{0.257} & 0.275 & 0.329 & 0.197 & 0.248 \\
 & 336 & \textcolor{blue}{0.242} & \textcolor{blue}{0.280} & \textcolor{blue}{0.242} & \textcolor{blue}{0.280} & 0.248 & 0.290 & 0.260 & 0.311 & 0.271 & 0.299 & 0.276 & 0.304 & 0.267 & 0.310 & 0.272 & 0.298 & 0.339 & 0.377 & \textcolor{red}{0.246} & \textcolor{red}{0.285} \\
 & 720 & \textcolor{red}{0.315} & \textcolor{blue}{0.332} & \textcolor{blue}{0.314} & \textcolor{red}{0.333} & 0.337 & 0.344 & 0.319 & 0.356 & 0.331 & 0.341 & 0.334 & 0.342 & 0.352 & 0.362 & 0.346 & 0.347 & 0.389 & 0.409 & 0.315 & 0.336 \\
\midrule
Traffic & 96 & \textcolor{red}{0.379} & \textcolor{red}{0.268} & \textcolor{blue}{0.361} & \textcolor{blue}{0.249} & 0.394 & 0.275 & 0.396 & 0.278 & 0.409 & 0.300 & 0.407 & 0.292 & 0.570 & 0.310 & 0.420 & 0.311 & 0.576 & 0.359 & 0.391 & 0.282 \\
 & 192 & \textcolor{red}{0.395} & 0.275 & \textcolor{blue}{0.379} & \textcolor{blue}{0.257} & 0.402 & 0.273 & 0.408 & 0.284 & 0.433 & \textcolor{red}{0.272} & 0.415 & 0.294 & 0.577 & 0.321 & 0.430 & 0.298 & 0.610 & 0.380 & 0.404 & 0.287 \\
 & 336 & \textcolor{red}{0.404} & \textcolor{red}{0.280} & \textcolor{blue}{0.392} & \textcolor{blue}{0.265} & 0.410 & 0.280 & 0.420 & 0.293 & 0.424 & 0.299 & 0.421 & 0.292 & 0.588 & 0.324 & 0.450 & 0.305 & 0.608 & 0.375 & 0.414 & 0.292 \\
 & 720 & \textcolor{red}{0.437} & \textcolor{red}{0.298} & \textcolor{blue}{0.432} & \textcolor{blue}{0.286} & 0.453 & 0.307 & 0.456 & 0.311 & 0.488 & 0.341 & 0.456 & 0.311 & 0.597 & 0.337 & 0.470 & 0.330 & 0.621 & 0.375 & 0.450 & 0.310 \\
\midrule
Electricity & 96 & \textcolor{red}{0.130} & \textcolor{red}{0.226} & \textcolor{blue}{0.129} & \textcolor{blue}{0.222} & 0.133 & 0.230 & 0.140 & 0.237 & 0.171 & 0.273 & 0.155 & 0.252 & 0.173 & 0.275 & 0.146 & 0.340 & 0.186 & 0.302 & 0.136 & 0.233 \\
 & 192 & \textcolor{red}{0.149} & \textcolor{red}{0.243} & 0.151 & 0.245 & \textcolor{blue}{0.147} & \textcolor{blue}{0.241} & 0.154 & 0.250 & 0.191 & 0.292 & 0.168 & 0.263 & 0.195 & 0.288 & 0.172 & 0.266 & 0.197 & 0.311 & 0.152 & 0.247 \\
 & 336 & 0.165 & \textcolor{blue}{0.260} & \textcolor{red}{0.164} & \textcolor{blue}{0.260} & \textcolor{blue}{0.162} & \textcolor{blue}{0.260} & 0.169 & \textcolor{red}{0.268} & 0.198 & 0.297 & 0.183 & 0.277 & 0.206 & 0.300 & 0.176 & 0.271 & 0.213 & 0.328 & 0.167 & 0.264 \\
 & 720 & 0.204 & 0.293 & \textcolor{red}{0.200} & \textcolor{red}{0.292} & \textcolor{blue}{0.192} & \textcolor{blue}{0.285} & 0.204 & 0.300 & 0.230 & 0.321 & 0.219 & 0.306 & 0.231 & 0.335 & 0.211 & 0.303 & 0.233 & 0.344 & 0.205 & 0.295 \\
 \midrule
ETTm1 & 96 & \textcolor{blue}{0.284} & \textcolor{blue}{0.338} & \textcolor{red}{0.290} & \textcolor{red}{0.344} & 0.310 & 0.356 & 0.304 & 0.348 & 0.327 & 0.363 & 0.329 & 0.363 & 0.335 & 0.373 & 0.324 & 0.367 & 0.326 & 0.390 & 0.293 & 0.349 \\
 & 192 & \textcolor{red}{0.327} & \textcolor{blue}{0.362} & 0.333 & 0.371 & 0.342 & 0.375 & 0.342 & 0.375 & 0.356 & 0.381 & 0.353 & 0.378 & 0.372 & 0.390 & 0.369 & 0.388 & 0.365 & 0.415 & \textcolor{blue}{0.326} & \textcolor{red}{0.368} \\
 & 336 & \textcolor{red}{0.360} & \textcolor{red}{0.388} & 0.370 & 0.392 & 0.370 & 0.389 & 0.369 & 0.389 & 0.387 & 0.397 & 0.382 & 0.394 & 0.403 & 0.411 & 0.396 & 0.408 & 0.392 & 0.425 & \textcolor{blue}{0.352} & \textcolor{blue}{0.384} \\
 & 720 & \textcolor{red}{0.414} & \textcolor{blue}{0.416} & 0.416 & 0.420 & 0.420 & \textcolor{red}{0.418} & 0.424 & 0.422 & 0.441 & 0.425 & 0.429 & \textcolor{red}{0.418} & 0.461 & 0.442 & 0.451 & 0.439 & 0.446 & 0.458 & \textcolor{blue}{0.405} & \textcolor{blue}{0.416} \\
\midrule
ETTm2 & 96 & \textcolor{blue}{0.163} & \textcolor{blue}{0.252} & \textcolor{red}{0.166} & \textcolor{red}{0.256} & 0.168 & 0.261 & 0.170 & 0.264 & 0.190 & 0.284 & 0.181 & 0.274 & 0.176 & 0.266 & 0.179 & 0.269 & 0.180 & 0.271 & 0.170 & 0.260 \\
 & 192 & \textcolor{blue}{0.217} & \textcolor{blue}{0.290} & \textcolor{red}{0.223} & \textcolor{red}{0.296} & 0.231 & 0.305 & 0.239 & 0.314 & 0.250 & 0.320 & 0.233 & 0.306 & 0.240 & 0.307 & 0.244 & 0.309 & 0.252 & 0.318 & 0.227 & 0.297 \\
 & 336 & \textcolor{blue}{0.266} & \textcolor{blue}{0.323} & 0.273 & 0.329 & \textcolor{red}{0.272} & \textcolor{red}{0.328} & 0.293 & 0.356 & 0.301 & 0.350 & 0.285 & 0.338 & 0.304 & 0.345 & 0.303 & 0.346 & 0.324 & 0.364 & 0.275 & 0.328 \\
 & 720 & \textcolor{blue}{0.346} & \textcolor{blue}{0.376} & \textcolor{red}{0.363} & \textcolor{red}{0.385} & 0.375 & 0.394 & 0.432 & 0.448 & 0.389 & 0.402 & 0.375 & 0.390 & 0.406 & 0.400 & 0.400 & 0.400 & 0.410 & 0.420 & 0.363 & 0.387 \\
\midrule
ETTh1 & 96 & \textcolor{blue}{0.360} & \textcolor{blue}{0.390} & 0.372 & 0.401 & 0.375 & 0.400 & \textcolor{red}{0.371} & 0.395 & 0.388 & 0.408 & 0.381 & 0.402 & 0.382 & 0.398 & 0.372 & \textcolor{red}{0.394} & 0.376 & 0.399 & 0.387 & 0.410 \\
 & 192 & \textcolor{blue}{0.396} & \textcolor{blue}{0.412} & 0.413 & 0.429 & 0.409 & 0.422 & \textcolor{red}{0.405} & \textcolor{red}{0.416} & 0.421 & 0.426 & 0.409 & 0.418 & 0.427 & 0.425 & 0.420 & 0.425 & 0.423 & 0.416 & 0.410 & 0.426 \\
 & 336 & \textcolor{blue}{0.420} & \textcolor{red}{0.429} & 0.434 & 0.446 & 0.438 & 0.439 & 0.466 & 0.469 & 0.430 & 0.434 & 0.423 & \textcolor{blue}{0.425} & 0.465 & 0.445 & 0.453 & 0.440 & 0.444 & 0.443 & \textcolor{red}{0.422} & 0.437 \\
 & 720 & 0.448 & 0.460 & 0.455 & 0.473 & 0.467 & 0.474 & 0.490 & 0.503 & \textcolor{red}{0.440} & \textcolor{red}{0.459} & \textcolor{blue}{0.427} & \textcolor{blue}{0.449} & 0.472 & 0.468 & 0.470 & 0.470 & 0.469 & 0.490 & 0.454 & 0.472 \\
\midrule
 ETTh2 & 96 & \textcolor{red}{0.274} & \textcolor{blue}{0.338} & 0.275 & \textcolor{red}{0.339} & \textcolor{blue}{0.271} & 0.339 & 0.290 & 0.357 & 0.305 & 0.367 & 0.295 & 0.358 & 0.309 & 0.359 & 0.293 & 0.340 & 0.332 & 0.374 & 0.288 & 0.345 \\
 & 192 & \textcolor{red}{0.337} & \textcolor{blue}{0.379} & 0.340 & \textcolor{red}{0.381} & \textcolor{blue}{0.330} & 0.384 & 0.377 & 0.419 & 0.350 & 0.393 & 0.340 & 0.386 & 0.390 & 0.406 & 0.375 & 0.390 & 0.407 & 0.446 & 0.349 & 0.386 \\
 & 336 & \textcolor{red}{0.360} & 0.410 & 0.364 & \textcolor{red}{0.402} & 0.365 & 0.411 & 0.487 & 0.486 & 0.360 & 0.404 & \textcolor{blue}{0.350} & \textcolor{blue}{0.395} & 0.465 & 0.445 & 0.378 & 0.406 & 0.400 & 0.447 & 0.369 & 0.408 \\
 & 720 & 0.398 & 0.434 & \textcolor{blue}{0.389} & \textcolor{red}{0.429} & 0.402 & 0.441 & 0.755 & 0.613 & 0.402 & 0.435 & \textcolor{red}{0.391} & \textcolor{blue}{0.428} & 0.472 & 0.468 & 0.409 & 0.432 & 0.412 & 0.469 & 0.403 & 0.439 \\
 \midrule
\textbf{First (Blue)} &  & \textbf{11} & \textbf{14} & \textbf{9} & \textbf{9} & \textbf{5} & \textbf{3} & \textbf{0} & \textbf{0} & \textbf{0} & \textbf{0} & \textbf{2} & \textbf{4} & \textbf{0} &\textbf{0}& \textbf{0} & \textbf{0} & \textbf{0} & \textbf{0} & \textbf{3} & \textbf{2} \\
\textbf{Second (Red)}  &  & \textbf{13} & \textbf{9} & \textbf{8} & \textbf{11} & \textbf{2} & \textbf{3} & \textbf{2} & \textbf{2} & \textbf{1} & \textbf{2} & \textbf{1} & \textbf{1} & \textbf{0}& \textbf{0} & \textbf{0} & \textbf{2} & \textbf{0} & \textbf{0} & \textbf{2} & \textbf{2}\\
\bottomrule
\end{tabular}
}
\end{table}

\begin{table}[ht]
\centering
\caption{Short-term forecasting performance (MSE and MAE) across models and PEMS datasets. Blue highlights denote the best score per row, and red indicates the second-best. We report performance for baselines as reported in ~\cite{liu2024itransformer}.}
\noindent\resizebox{\textwidth}{!}{%
\label{tab:performance_short}
\begin{tabular}{llcccccccccccccccccccc}
\toprule
\multirow{2}{*}{\textbf{}} & \multirow{2}{*}{\textbf{Horizon}} & \multicolumn{2}{c}{\textbf{FaCTR}} & \multicolumn{2}{c}{\textbf{iTransformer}} & \multicolumn{2}{c}{\textbf{SCINet}} & \multicolumn{2}{c}{\textbf{Rlinear}} & \multicolumn{2}{c}{\textbf{PatchTST}} & \multicolumn{2}{c}{\textbf{CrossFormer}} & \multicolumn{2}{c}{\textbf{TimesNet}} & \multicolumn{2}{c}{\textbf{Dlinear}} & \multicolumn{2}{c}{\textbf{FEDFormer}} & \multicolumn{2}{c}{\textbf{Autoformer}} \\
\cmidrule(lr){3-4} \cmidrule(lr){5-6} \cmidrule(lr){7-8} \cmidrule(lr){9-10} \cmidrule(lr){11-12} \cmidrule(lr){13-14} \cmidrule(lr){15-16} \cmidrule(lr){17-18} \cmidrule(lr){19-20} \cmidrule(lr){21-22} 
& & MSE & MAE & MSE & MAE & MSE & MAE & MSE & MAE & MSE & MAE & MSE & MAE & MSE & MAE & MSE & MAE & MSE & MAE & MSE & MAE \\ \midrule
PEMS03 & 
 12 & \textcolor{blue}{0.064} & \textcolor{blue}{0.167} & 0.071 & 0.174 & \textcolor{red}{0.066} & \textcolor{red}{0.172} & 0.126 & 0.236 & 0.099 & 0.216 & 0.090 & 0.203 & 0.085 & 0.192 & 0.122 & 0.243 & 0.126 & 0.251 & 0.272 & 0.385 \\
& 24 & \textcolor{blue}{0.083} & \textcolor{blue}{0.189} & 0.093 & 0.201 & \textcolor{red}{0.085} & \textcolor{red}{0.198} & 0.246 & 0.334 & 0.142 & 0.259 & 0.121 & 0.240 & 0.118 & 0.223 & 0.201 & 0.317 & 0.149 & 0.275 & 0.334 & 0.440 \\
& 48 & \textcolor{blue}{0.110} & \textcolor{blue}{0.217} & \textcolor{red}{0.125} &\textcolor{red}{0.236} & 0.127 & 0.238 & 0.551 & 0.529 & 0.211 & 0.319 & 0.202 & 0.317 & 0.155 & 0.260 & 0.333 & 0.425 & 0.227 & 0.348 & 1.032 & 0.782 \\
& 96 & \textcolor{blue}{0.137} & \textcolor{blue}{0.243} & \textcolor{red}{0.164} & \textcolor{red}{0.275} & 0.178 & 0.287 & 1.057 & 0.787 & 0.269 & 0.370 & 0.262 & 0.367 & 0.228 & 0.317 & 0.457 & 0.515 & 0.348 & 0.434 & 1.031 & 0.796 \\
\midrule
PEMS04 & 12 & \textcolor{red}{0.078} & \textcolor{red}{0.183} & \textcolor{red}{0.078} & \textcolor{red}{0.183} & \textcolor{blue}{0.073} & \textcolor{blue}{0.177} & 0.138 & 0.252 & 0.105 & 0.224 & 0.098 & 0.218 & 0.087 & 0.195 & 0.148 & 0.272 & 0.138 & 0.262 & 0.424 & 0.491 \\
 & 24 & \textcolor{red}{0.093} & \textcolor{red}{0.201} & 0.095 & 0.205 & \textcolor{blue}{0.084} & \textcolor{blue}{0.193} & 0.258 & 0.348 & 0.153 & 0.275 & 0.131 & 0.256 & 0.103 & 0.215 & 0.224 & 0.340 & 0.177 & 0.293 & 0.459 & 0.509 \\
 & 48 & \textcolor{red}{0.118} & \textcolor{red}{0.229} & 0.120 & 0.233 & \textcolor{blue}{0.099} & \textcolor{blue}{0.211} & 0.572 & 0.544 & 0.229 & 0.339 & 0.205 & 0.326 & 0.136 & 0.250 & 0.355 & 0.437 & 0.270 & 0.368 & 0.646 & 0.610 \\
 & 96 & \textcolor{red}{0.140} & \textcolor{red}{0.250} & 0.150 & 0.262 & \textcolor{blue}{0.114} & \textcolor{blue}{0.227} & 1.137 & 0.820 & 0.291 & 0.389 & 0.402 & 0.457 & 0.190 & 0.303 & 0.452 & 0.504 & 0.341 & 0.427 & 0.912 & 0.748 \\
\midrule
PEMS07 & 12 & \textcolor{blue}{0.056} & \textcolor{blue}{0.154} & \textcolor{red}{0.067} & \textcolor{red}{0.165} & 0.068 & 0.171 & 0.118 & 0.235 & 0.095 & 0.207 & 0.094 & 0.200 & 0.082 & 0.181 & 0.115 & 0.242 & 0.109 & 0.225 & 0.199 & 0.336 \\
 & 24 & \textcolor{blue}{0.071} & \textcolor{blue}{0.175} & \textcolor{red}{0.088} & \textcolor{red}{0.190} & 0.119 & 0.225 & 0.242 & 0.341 & 0.150 & 0.262 & 0.139 & 0.247 & 0.101 & 0.204 & 0.210 & 0.329 & 0.125 & 0.244 & 0.323 & 0.420 \\
 & 48 & \textcolor{blue}{0.096} & \textcolor{blue}{0.204} & \textcolor{red}{0.110} & \textcolor{red}{0.215} & 0.149 & 0.237 & 0.562 & 0.541 & 0.253 & 0.340 & 0.311 & 0.369 & 0.134 & 0.238 & 0.398 & 0.458 & 0.165 & 0.288 & 0.390 & 0.470 \\
 & 96 & \textcolor{blue}{0.118} & \textcolor{blue}{0.226} & \textcolor{red}{0.139} & \textcolor{red}{0.245} & 0.141 & 0.234 & 1.096 & 0.795 & 0.346 & 0.404 & 0.396 & 0.442 & 0.181 & 0.279 & 0.594 & 0.553 & 0.262 & 0.376 & 0.554 & 0.578 \\
 \midrule
PEMS08 & 12 & \textcolor{red}{0.083} & 0.185 & \textcolor{blue}{0.079} & \textcolor{blue}{0.182} & 0.087 & \textcolor{red}{0.184} & 0.133 & 0.247 & 0.168 & 0.232 & 0.165 & 0.214 & 0.112 & 0.212 & 0.154 & 0.276 & 0.173 & 0.273 & 0.436 & 0.485 \\
 & 24 & \textcolor{blue}{0.104} & \textcolor{blue}{0.203} & \textcolor{red}{0.115} & \textcolor{red}{0.219} & 0.122 & 0.221 & 0.249 & 0.343 & 0.224 & 0.281 & 0.215 & 0.260 & 0.141 & 0.238 & 0.248 & 0.353 & 0.210 & 0.301 & 0.467 & 0.502 \\
 & 48 & \textcolor{blue}{0.157} & \textcolor{blue}{0.235} & \textcolor{red}{0.186} & \textcolor{blue}{0.235} & 0.189 & \textcolor{red}{0.270} & 0.569 & 0.544 & 0.321 & 0.354 & 0.315 & 0.355 & 0.198 & 0.283 & 0.440 & 0.470 & 0.320 & 0.394 & 0.966 & 0.733 \\
 & 96 & \textcolor{blue}{0.213} & \textcolor{blue}{0.260} & \textcolor{red}{0.221} & \textcolor{red}{0.267} & 0.236 & 0.300 & 1.166 & 0.814 & 0.408 & 0.417 & 0.377 & 0.397 & 0.320 & 0.351 & 0.674 & 0.565 & 0.442 & 0.465 & 1.385 & 0.915 \\
 \midrule
 \textbf{First (Blue)} &  & \textbf{11} & \textbf{11} & \textbf{1} & \textbf{1} & \textbf{4} & \textbf{4} & \textbf{0} & \textbf{0} & \textbf{0} & \textbf{0} & \textbf{0} & \textbf{0} & \textbf{0} &\textbf{0}& \textbf{0} & \textbf{0} & \textbf{0} & \textbf{0} & \textbf{0} & \textbf{0} \\
\textbf{Second (Red)}  &  & \textbf{5} & \textbf{5} & \textbf{9} & \textbf{10} & \textbf{2} & \textbf{4} & \textbf{0} & \textbf{0} & \textbf{0} & \textbf{0} & \textbf{0} & \textbf{0} & \textbf{0}& \textbf{0} & \textbf{0} & \textbf{0} & \textbf{0} & \textbf{0} & \textbf{0} & \textbf{0}\\
\bottomrule
\end{tabular}
}
\end{table}

\subsection{Interpretability via a Retail Demand Forecasting Case Study}
To probe interpretability, we construct a synthetic multivariate demand forecasting dataset spanning multiple years and 8 channels, where each channel simulates a distinct, interpretable demand pattern commonly observed in real-world settings. These include seasonal baselines (C1–C3), with C1 and C2 being perfectly redundant; trend-overlaid seasonality (C4); pure noise (C5); direct promotion triggers (C6); a positively lagged response to promotions (C7); and seasonality modulated by negative promotional effects (C8, i.e., cannibalization). This controlled setup enables precise diagnosis of forecasting behavior under structured cross-channel dependencies (see Appendix~\ref{appendix:casestudy} for details).

\begin{figure}[h]
    \centering
    \begin{minipage}{0.52\linewidth}
        \centering
        \includegraphics[width=\linewidth]{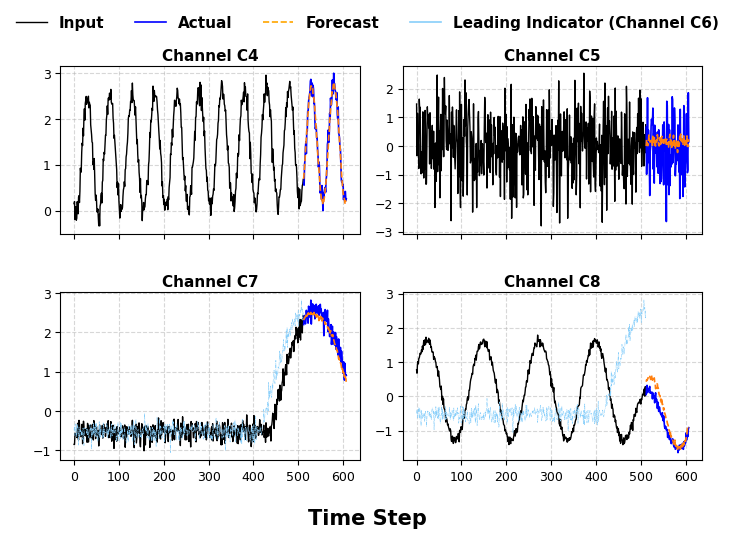}
        \captionof{figure}{Forecasts of sequence 240}
        \label{fig:forecasts}
    \end{minipage}
    \begin{minipage}{0.40\linewidth}
        \centering
        \includegraphics[width=\linewidth]{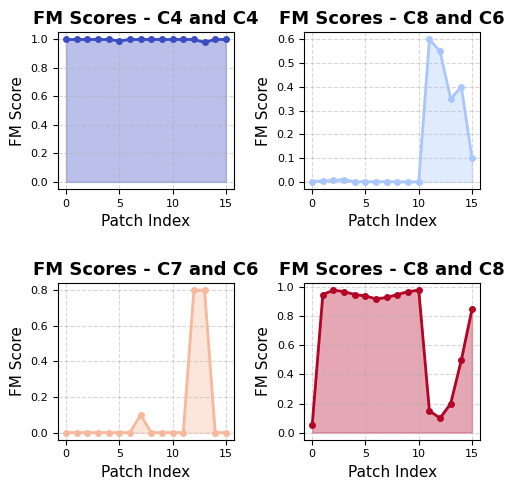}
        \captionof{figure}{FM Scores of sequence 240}
        \label{fig:interpetplots}
    \end{minipage}
\end{figure}

As shown in Figure~\ref{fig:forecasts}, FaCTR accurately forecasts seasonal and trending patterns (C4), avoids overfitting noise (C5), and captures meaningful cross-channel dependencies. Notably, C7 rises after a delayed response to the promotion signal in C6—despite promotions occurring at irregular intervals—indicating learned causal lag. In C8, the seasonal baseline is correctly suppressed during C6 peaks while preserving intrinsic seasonality, demonstrating retention of temporal dynamics.

To further assess interpretability, we analyze FM scores across channels (Figure~\ref{fig:interpetplots}). For C4, the model relies entirely on autoregressive signals, as expected for a clean seasonal-trend pattern. For C7, we observe sharp FM attention to C6 around patch index 13, where the promotion signal in C6 begins to peak. This alignment confirms learned causal influence: C6 triggers C7’s rise and informs its decline. Similarly, C8 predominantly attends to itself, consistent with its intrinsic seasonality. However, when C6 peaks, FM scores at patches 12–13 show a temporary shift in C8’s attention toward C6—signaling the release of promotion-induced suppression and a return to its seasonal cycle. These patterns highlight FaCTR’s ability to capture both temporal and cross-channel causality, supporting its value for transparent, structured forecasting.

\subsection{Representation Learning}

\begin{wraptable}[12]{h}{0.55\textwidth}
\vspace{-1.5ex}
\centering
\caption{Forecasting performance on ETTh2 across varying prediction lengths using linear-probed, fine-tuned, and from-scratch supervised baseline versions of FaCTR. Lower MSE and MAE are better. Bold indicates best performance.}
\label{tab:pretrain}
\resizebox{0.55\textwidth}{!}{%
\begin{tabular}{c|cc|cc|cc}
\toprule
\multirow{2}{*}{\textbf{Horizon}} &
\multicolumn{2}{c|}{\textbf{Linear Probing}} &
\multicolumn{2}{c|}{\textbf{Fine-Tuning}} &
\multicolumn{2}{c}{\textbf{Trained from Scratch}} \\
 & \textbf{MSE} & \textbf{MAE} & \textbf{MSE} & \textbf{MAE} & \textbf{MSE} & \textbf{MAE} \\
\midrule
96   & 0.286 & 0.346 & 0.283 & 0.341 & \textbf{0.274} & \textbf{0.338} \\
192  & 0.344 & 0.384 & 0.345 & 0.384 & \textbf{0.337} & \textbf{0.379} \\
336  & 0.366 & 0.409 & 0.368 & \textbf{0.407} & \textbf{0.360} & 0.410 \\
720  & \textbf{0.394} & \textbf{0.434} & 0.399 & 0.436 & 0.398 & \textbf{0.434} \\
\bottomrule
\end{tabular}
}
\vspace{-1ex}
\end{wraptable}

We assess the benefits of self-supervised representation learning by pretraining FaCTR on the ETTh1 dataset using a random patch-masked reconstruction objective. Specifically, we mask 45\% of the non-overlapping input patches with zeros and train the model to reconstruct the original sequence. This encourages the model to learn temporal dependencies and cross-channel structure that generalize beyond the training data. Pretraining is conducted for 100 epochs using reconstruction loss as the objective (see Appendix~\ref{appendix:hyparams} for details). Once pretrained, we evaluate the learned representations by transferring the model to a supervised forecasting task on ETTh2. We consider two transfer strategies:  
(a) \textit{Linear probing}, where the pretrained backbone is frozen and only the prediction head is trained;  
(b) \textit{Full fine-tuning}, where the entire model is updated end-to-end.

As a baseline, we also evaluate a version trained from scratch on ETTh2 without pretraining. Table~\ref{tab:pretrain} shows that self-supervised pretraining improves performance across all horizons. Linear probing remains consistently competitive and achieves the best MSE at the longest horizon (720), underscoring the strength of the learned representations. Full fine-tuning further improves accuracy at shorter horizons, while the non-pretrained model remains the overall top performer on average. These results demonstrate that patch-masked pretraining enables efficient and transferable spatiotemporal representations for downstream forecasting. See Appendix ~\ref{appendix:replearn} for a more detailed analysis on the setup and results.

\subsection{Ablation Study}
To isolate the contribution of each structural component in FaCTR, we conduct an ablation study on three datasets (ETTh1, ETTh2, and Weather) across all four prediction horizons. We evaluate three variants: (i) \textbf{Temporal Only} — a baseline using channel-independent vanilla temporal attention with patching; (ii) \textbf{+FM} — which augments the baseline with FM-based spatial attention and gated temporal–spatial fusion; and (iii) \textbf{Full FaCTR} — which further incorporates embedding-wise mixing. All variants share the same projection head and training setup.
\begin{wraptable}[15]{h}{0.50\textwidth}
\vspace{-1.5ex}
\centering
\caption{Performance comparison of different FaCTR configurations on multiple datasets.}
\label{tab:ablation}
\resizebox{0.50\textwidth}{!}{%
\begin{tabular}{llcccccc}
\toprule
\textbf{Dataset} & \textbf{Horizon} & \multicolumn{2}{c}{Temporal Only} & \multicolumn{2}{c}{+FM} & \multicolumn{2}{c}{Full FaCTR} \\
\cmidrule(lr){3-4} \cmidrule(lr){5-6} \cmidrule(lr){7-8}
& & MSE & MAE & MSE & MAE & MSE & MAE \\
\midrule
\multirow{4}{*}{Weather}
& 96  & 0.173 & 0.223 & 0.165 & 0.218 & \textbf{0.148} & \textbf{0.200} \\
& 192 & 0.221 & 0.267 & 0.209 & 0.256 & \textbf{0.191} & \textbf{0.240} \\
& 336 & 0.261 & 0.295 & 0.258 & 0.292 & \textbf{0.242} & \textbf{0.280} \\
& 720 & 0.323 & 0.338 & 0.318 & 0.334 & \textbf{0.315} & \textbf{0.332} \\
\midrule
\multirow{4}{*}{ETTh2}
& 96  & 0.364 & 0.414 & 0.289 & 0.347 & \textbf{0.274} & \textbf{0.338} \\
& 192 & 0.345 & 0.382 & 0.360 & 0.391 & \textbf{0.337} & \textbf{0.379} \\
& 336 & 0.370 & 0.409 & 0.369 & 0.408 & \textbf{0.360} & \textbf{0.408} \\
& 720 & 0.401 & 0.440 & 0.400 & 0.436 & \textbf{0.398} & \textbf{0.434} \\
\midrule
\multirow{4}{*}{ETTh1}
& 96  & 0.365 & 0.394 & 0.362 & 0.393 & \textbf{0.360} & \textbf{0.391} \\
& 192 & 0.399 & 0.416 & 0.396 & 0.414 & \textbf{0.395} & \textbf{0.413} \\
& 336 & 0.425 & 0.435 & \textbf{0.422} & 0.431 & \textbf{0.422} & \textbf{0.430} \\
& 720 & 0.450 & 0.467 & 0.450 & 0.464 & \textbf{0.447} & \textbf{0.462} \\
\bottomrule
\end{tabular}
}
\vspace{-1ex}
\end{wraptable}
 Results in Table~\ref{tab:ablation} show that spatial reasoning via FM consistently improves performance over temporal-only models, and that Full FaCTR achieves the lowest error across nearly all settings. These gains are especially pronounced at shorter forecast horizons, where fine-grained temporal and cross-channel cues are more predictive; in contrast, longer horizons are inherently noisier and more stochastic, making gains harder to realize. This highlights the complementary nature of temporal modeling, FM-based spatial reasoning, and embedding mixing in enabling FaCTR to generalize robustly across domains.

\section{Conclusion and Future Work}
We presented FaCTR, an interpretable structured spatiotemporal Transformer for multivariate time series forecasting that introduces inductive priors via patch-based temporal modeling and low-rank spatial factorization. By conditioning on rich spatial and dynamic context, it generalizes well across benchmarks while maintaining a compact parameter footprint. FaCTR supports exogenous conditioning, temporal fusion, and adaptive channel mixing within a unified and interpretable framework capable of representation learning. Future directions include extending support for future-known covariates and hierarchical reconciliation across spatial levels.

\begin{small}

\end{small}

\newpage
\appendix

\section{Proofs}
\subsection{Breakdown of Univariate Dependency Assumptions}
\label{appendix:univariate}

In language, token sequences follow autoregressive chains where token likelihood is well-approximated by a small context window, consistent with the Markov property:

\begin{equation}
    P(x_t \mid x_{<t}) \approx P(x_t \mid x_{t-k:t-1}) \quad \text{for small } k\ \text{\cite{dai2019transformerxl}}
\end{equation}

In vision, spatial locality enables attention factorization, where the mutual information between distant patches is negligible:
\begin{equation}
    I(p_{ij}; p_{kl}) \approx 0 \quad \text{when} \quad \|(i,j)-(k,l)\| > r\ \text{\cite{cover1981}}
\end{equation}

Time series data violate both assumptions. A scalar observation like temperature $y_t$ carries low intrinsic entropy~\cite{hyndman2021}—$H(y_t) \ll H(x_{\text{token}})$—and must be contextualized using both its \textbf{temporal history} ($y_{<t}$) and \textbf{cross-variable context} ($z_t$). This is quantified via conditional entropy and necessitates conditioning with exogenous signals~\cite{granger1988causality}:

\begin{equation}
    H(y_t \mid y_{<t}, z_t) \ll H(y_t \mid y_{<t})
\end{equation}

\subsection{Cross-Channel Time series inductive biases}
\label{appendix:inductivebias}

\paragraph{Timescale Separation.}
In many weakly coupled systems, cross-channel dependencies decay with time difference $\tau$ much faster than temporal autocorrelations. Formally, the mutual information between variables decays exponentially:
    \[
        I(y_t^{(i)}; y_{t+\tau}^{(j)}) \leq \alpha e^{-\beta|\tau|} \quad \text{(Theorem 3, \cite{cover1981})}
    \]
    where $\alpha, \beta$ are system constants. This suggests temporal modeling requires fine-grained attention across $\tau$, while cross-channel effects can be modeled using coarse temporal aggregations (e.g., hourly or daily bins).

    \textit{Example: Power Grid Dynamics} — In frequency-voltage coupling:
    \[
        \Delta f_t = \sum_{j=1}^C K_{ij} \Delta V_{t - \tau_{ij}}
    \]
    where $\tau_{ij}$ spans milliseconds to minutes depending on transmission distance, but the dependency evolves slowly and predictably.\\

\paragraph{Structural Invariance.}
Many cross-channel relationships are not only slow-changing, but structurally invariant over time. Examples include physical laws, conservation constraints, and stable inter-variable couplings. These relationships can often be captured by fixed matrices, embeddings, or low-rank factorization.

    \textit{Example: Structural Causality in Retail Demand} — Cross-price elasticities between products are governed by market structure, not transient dynamics:
    \[
        \frac{\partial \text{Sales}_t^{(i)}}{\partial \text{Price}_t^{(j)}} = 
        \begin{cases}
            \gamma_{ij} & \text{if } j \in \text{competitors} \\
            0 & \text{otherwise}
        \end{cases}
    \]
    Such relationships remain stable over weeks, while intra-product sales autocorrelate on minute-level resolution~\cite{hanssens2015}. Failing to exploit this separation burdens attention mechanisms with relearning static structure repeatedly.

\subsection{Mutual Information Sparsity in Multivariate Time Series}
\label{appendix:mi-sparsity}

Let $x_t^{(i)}$ and $x_t^{(j)}$ denote time-aligned observations from channels $i$ and $j$ at time $t$. Empirical studies across domains such as climate science, energy, and retail show that the mutual information between most variable pairs is negligible:
\[
I(x_t^{(i)}; x_t^{(j)}) < \varepsilon, \quad \forall (i,j) \notin \mathcal{E}, \quad \mathcal{E} \subseteq [C] \times [C],
\]
for some small $\varepsilon > 0$ and sparse edge set $\mathcal{E}$. This reflects a structural prior that most channels are conditionally independent or weakly coupled—justifying low-rank and sparsity-aware modeling approaches such as the Factorization Machine.

\section{Model Architecture Design Rationale}
\label{appendix:designrat}
\subsection{Embedding Module}
A key limitation of purely autoregressive forecasting models is their reliance on past values alone to make future predictions. In real-world settings such as retail, this assumption is often inadequate: demand is heavily influenced by structured contextual signals like calendar information (e.g., day-of-week, holidays), promotional activity, product characteristics, and store-level metadata.

By integrating these signals at the very beginning of the architecture, we inject structural inductive biases into the model — guiding it to condition its predictions not just on past values, but also on external drivers and context. This helps the model generalize better, especially in cold-start scenarios, sparse regimes, or periods with external shocks (e.g., a new promotion or holiday).

In particular, we explicitly incorporate static and categorical covariates through learned embeddings, enabling the model to capture complex entity-specific or context-specific behavior patterns. This approach alleviates the need for extensive manual feature engineering, allowing the model to automatically learn useful representations from raw metadata — a critical advantage in large-scale or heterogeneous forecasting environments.

This design aligns with the broader trend in time series forecasting toward context-aware representation learning, and serves as a foundation for the downstream attention and spatial reasoning layers.

\subsection{Temporal Attention Module}

While contextual embeddings provide rich auxiliary signals, capturing temporal dependencies is foundational to time series forecasting. Time series often exhibit strong autocorrelation, seasonal cycles, and trend components — patterns that evolve over time and are critical to accurate prediction.

By performing self-attention independently for each time series channel (i.e., product or store), the model learns per-channel temporal structure, adapting to diverse dynamics without parameter sharing across channels. This design encourages modularity and interpretability, since each product’s temporal patterns are modeled in isolation before any cross-channel interaction occurs.

\subsection{Factorization Machine and Gating Module}
While temporal attention enables each channel to independently learn rich temporal structure, it does not capture inter-series (cross-channel) dependencies. In many real-world forecasting problems, channels are not independent — demand for one product or store can influence, or be influenced by, others. The Factorization Machine (FM) Module is designed to model such spatial relationships between channels, while preserving computational efficiency and structural modularity.

We highlight three key components of this design:

\textbf{Low-Rank Spatial Projection}
The output of temporal attention lies in a temporally contextualized latent space. However, to model cross-channel dependencies, we must operate in a space where channels can be compared and interacted. To achieve this, we project the temporal output into a spatial representation space that is compatible with the FM-based similarity scores.

This mirrors the core design in traditional attention mechanisms, where inputs are transformed into query (Q), key (K), and value (V) spaces using learned projections ($W_{Q}$, $W_{K}$, $W_{V}$) to enable meaningful interactions. In our case, the FM module serves as a lightweight substitute for the Q-K dot product — computing similarity scores without separate Q and K projections. However, in order to apply these FM-derived attention scores effectively, we must project the temporal attention output into a "value" space that is structurally aligned with the similarity scores.

We achieve this through a low-rank transformation, which learns compact, abstract representations of temporal features that are optimized for spatial reasoning. This approach not only reduces computational cost but also introduces an implicit regularization effect — encouraging the model to learn disentangled and generalizable inter-series dependencies.

\textbf{FM Score Computation as Structural Attention}
Traditional attention mechanisms compute scores using learned query-key dot products. In contrast, our FM-style interaction replaces the notion of explicit queries and keys with a learned structural context derived from the covariate-enriched embeddings. 

This means we compute attention scores between channel pairs using an inner product over the low-rank projected contextual representations. These scores act as data-driven similarity measures, capturing how strongly one channel (e.g., a product or store) relates to another based on its combined temporal and structural context. This FM-inspired score matrix allows the model to dynamically decide which other channels to attend to, based on their learned relevance rather than fixed identity. 

By doing so, we move beyond simple dot-product attention and enable structural attention, where interactions are governed by rich representations of the input space itself.

\textbf{Gating for Dynamic Residual Control}
Residual connections are a common tool in deep architectures, but blindly adding residuals may dilute useful signals or lead to unstable gradients — especially when combining signals from different modalities (temporal and spatial in this case). To address this, we introduce a learned gating mechanism that adaptively weighs the contribution of the temporal and spatial pathways. This enables the model to adaptively prioritize the more informative signal — for example (1) Relying more on temporal patterns for stable, high-volume products or (2) Leveraging spatial similarity for new or volatile items where peer behavior provides useful guidance.

The gating mechanism thus offers a principled alternative to static residuals, allowing the model to learn how to mix temporal and structural signals based on the underlying data characteristics.

\subsection{Embedding-wise Mixing and Projection head Module}

After modeling temporal dependencies via attention and cross-channel interactions via the factorization machine module, one key dimension remains: the feature (embedding) dimension. Time series often encode complex signals where dependencies exist not just across time or channels, but also within the latent representation space itself — e.g., interactions between learned seasonal patterns, event signals, and channel identity.
By operating across the D-dimensional embedding space, the MLP allows the model to learn cross-feature correlations. While temporal attention and FM capture dependencies over time and across series, feature mixing expands the effective receptive field (ERF) into the representation space. This ensures the model can capture global interactions across all three axes: time, channel, and features.

\section{Model and Experiment Details}

\subsection{Model details}

\subsubsection{Reversible Instance Normalization}
\label{appendix:revin}

To address non-stationarity and scale shifts in multivariate series, FaCTR employs \textbf{Reversible Instance Normalization (RevIN)} [38] applied independently to each channel. This approach ensures stable learning by normalizing each channel's time series and later restoring the original scale of the predictions.

\[
\text{RevIN}(x^{(k)}) = \gamma_k \cdot \frac{x^{(k)} - \mu^{(k)}}{\sigma^{(k)} + \epsilon} + \beta_k, \quad \text{for } k = 1, \ldots, C
\]

Here, \(\epsilon\) is a small constant added for numerical stability to prevent division by zero. \(\beta \in \mathbb{R}^C\) denotes the shift parameters used in normalization, and \(\gamma \in \mathbb{R}^C\) represents the learnable scale parameters used in the affine transformation.

\subsubsection{Sharpness Aware Minimization}
\label{appendix:sam}
Sharpness-Aware Minimization (SAM) is an optimization technique designed to improve model generalization by encouraging convergence to flat minima in the loss landscape. Unlike standard gradient descent, which minimizes loss at a single point in parameter space, SAM optimizes for parameters that perform well not only at the current point but also in a neighborhood around it. The result is a model that is less sensitive to parameter variations, which typically leads to improved robustness and generalization on unseen data. Please refer to ~\cite{ilbert2024} for more detailed proofs of the theorems associated.

\subsubsection{Time Complexity Breakdown}
\label{appendix:time-complexity}

Table~\ref{tab:time-complexity} provides a detailed breakdown of the time complexity for each major component of FaCTR. Let $B$ be the batch size, $C$ the number of channels, $N$ the number of temporal patches, $D$ the embedding dimension, $r$ the FM and spatial rank, and $T$ the forecast horizon.

\begin{table}[h]
\caption{Component-wise time complexity of FaCTR. $B$ and $D$ are omitted in the main paper for clarity.}
\label{tab:time-complexity}
\centering
\begin{tabular}{llc}
\toprule
\textbf{Component} & \textbf{Description} & \textbf{Time Complexity} \\
\midrule
Temporal Attention & Self-attention across $N$ per channel & $\mathcal{O}(B C N^2 D)$ \\
FM Similarity Matrix & Cross-channel dot products per patch & $\mathcal{O}(B C^2 N D)$ \\
Low-Rank Projection & Two-layer MLP for FM alignment & $\mathcal{O}(B C N D r)$ \\
Gated Fusion & Convex blend of temporal and spatial signals & $\mathcal{O}(B C N D^2)$ \\
Embedding-wise MLP & Pointwise FFN per $(b, c, n)$ & $\mathcal{O}(B C N D^2)$ \\
Forecast Projection & Linear map to forecast horizon $T$ & $\mathcal{O}(B C N D T)$ \\
\bottomrule
\end{tabular}
\end{table}

Omitting $B$ and $D$ for readability, the total time complexity (including the forecast projection term) is:
\[
\mathcal{O}(C N^2 + C^2 N + N T).
\]
This reflects contributions from temporal attention, FM-based cross-channel interaction, and the final projection head. In retail forecasting regimes—where $C$ can be in the thousands but $N$ and $T$ are typically small (e.g., $N = 8$, $T = 14$)—the overall runtime is dominated by the FM component and simplifies to $\mathcal{O}(C^2)$. Compared to full spatiotemporal attention, which scales as $\mathcal{O}((C N)^2)$, FaCTR achieves significantly better efficiency without sacrificing full coverage of temporal and cross-channel dependencies.

\subsection{Experiment details}
\label{appendix:expdet}
\subsubsection{Hardware}
\label{appendix:hardware}

All experiments are conducted on a \textbf{single NVIDIA A100 80GB GPU} using PyTorch. 

\subsubsection{Dataset details}
\label{appendix:datasets}

We conduct our evaluations on eight widely used, publicly available multivariate time series datasets for long-term forecasting and four PEMS datasets for short-term forecasting. The ETT datasets (ETTh1, ETTh2, ETTm1, ETTm2) comprise electricity transformer load readings collected from 2016 to 2018. The Electricity dataset captures power consumption patterns of 321 clients between 2012 and 2014. Traffic contains road occupancy rates from 862 sensors measured between 2015 and 2016. Weather includes 21 meteorological variables recorded throughout 2020. The  PEMS dataset consists of traffic data in California that was introduced in ~\cite{liu2022scinet}. For long-term forecasting datasets, the forecasting horizons are set to 96, 192, 336, and 720 time steps. For short-term forecasting, the forecasting horizons are set to 12, 24, 48 and 96. Table~\ref{tab:dataset_stats1} provides a summary of their key characteristics.

All datasets are chronologically split into training, validation, and test sets, following standard practice in recent time series forecasting benchmarks \cite{nie2023patchtst}. We use a context window of 512 for all prediction lengths. For the long-term forecasting datasets (ETTh1, ETTh2, ETTm1, ETTm2), we adopt a 12 months, 4 months, 4 months ratio for training, validation and testing respectively. The Electricity, Traffic, and Weather datasets are split using a 7:1:2 ratio for training, validation, and testing, respectively, while the short-term forecasting datasets (PEMS) follow a 6:2:2 ratio. These splits are applied chronologically to preserve temporal order and avoid future data leakage. All splits are based on predefined ratios, as commonly used in recent time series forecasting literature, ensuring fair and consistent evaluation across benchmarks.

\begin{table}[h]
\centering
\begin{tabular}{lcccccc}
\toprule
\textbf{Dataset} & \textbf{ETTh1/ETTh2} & \textbf{ETTm1/ETTm2} & \textbf{Electricity} & \textbf{Traffic} & \textbf{Weather} \\
\midrule
\# features       & 7     & 7      & 321    & 862   & 21 \\
Granularity       & 1 hour & 15 minutes & 1 hour & 1 hour & 10 minutes \\
Splits & 12/4/4 & 12/4/4 & 7:1:2 & 7:1:2 & 7:1:2 \\
\bottomrule
\end{tabular}
\\
\caption{Statistics of benchmark time series datasets for long-term time series forecasting.}
\label{tab:dataset_stats1}
\end{table}

\begin{table}[h]
\centering
\begin{tabular}{lccccc}
\toprule
\textbf{Dataset} & \textbf{PEMS03} & \textbf{PEMS04} & \textbf{PEMS07} & \textbf{PEMS08} \\
\midrule
\# features       & 358     & 307      & 883    & 170   \\
Granularity       & 5 minutes & 5 minutes & 5 minutes &  5 minutes\\
Splits & 6:2:2 & 6:2:2 & 6:2:2 & 6:2:2 \\
\bottomrule
\end{tabular}
\\
\caption{Statistics of benchmark time series datasets for short-term time series forecasting.}
\label{tab:dataset_stats}
\end{table}

\subsubsection{Hyperparameter Optimization}
\label{appendix:hyparams}

We train all models using a batch size of 32 for the ETT datasets and 64 for the larger datasets such as Electricity, Traffic, Weather and PEMS. A comprehensive hyperparameter search was conducted over several learning rates {1e-3, 5e-4, 1e-4} and learning rate schedulers, including CosineAnnealing, StepLR, and ReduceLROnPlateau. Based on validation performance, we selected a learning rate of 1e-4 in conjunction with cosine annealing with warm restarts as the default configuration.

We further experimented with various optimizers—Adam, AdamW, and SGD—and observed that Adam consistently outperformed others, particularly when combined with Sharpness-Aware Minimization (SAM). All models are trained for 150 epochs, with early stopping triggered after 10 epochs of no improvement in validation loss. All hyperparameter tuning was performed using the validation dataset.

To determine the optimal context window, we evaluated several lookback lengths—96, 336, 512, and 720—and found that a lookback of 512 offers the best trade-off between accuracy and efficiency, aligning well with the lightweight nature of our transformer-based architecture.

In our patch-based representation, we experimented with multiple combinations of patch lengths and strides. We adopt a non-overlapping patching scheme with both patch length and stride set to 32. For the Factorization Machine (FM) layer, we fix the rank, spatial rank and hidden dimension to 8, 8 and 32 respectively across all datasets, after running multiple ablations across dimensions - 8, 16, 32 and 64 for all the three.

We use the mean squared error (MSE) loss function for training all models. For representation learning, we use mean squared error as the reconstruction loss.

One of the key hyperparameters we tuned was the neighborhood size ($\rho$) used in SAM optimization. We conducted a thorough sweep by training 20 models for each prediction horizon, varying $\rho$ in the range [0.05, 1.0] with increments of 0.05. To ensure the robustness of our findings, each configuration was run multiple times. Table~\ref{tab:rho_values} lists the neighborhood sizes $\rho$ used for each dataset for FaCTR and other SAM-based models (SAMformer and TSMixer + SAM). These values control the radius for sharpness-aware updates during optimization, with tuning done via grid search on the validation set.

\begin{table}[h]
\centering
\caption{Neighborhood size $\rho$ used in SAM-based models across different datasets and prediction horizons for long-term forecasting datasets.}
\label{tab:rho_values}
\begin{tabular}{llcccccccc}
\toprule
\textbf{Model} & \textbf{Horizon} & \textbf{ETTh1} & \textbf{ETTh2} & \textbf{ETTm1} & \textbf{ETTm2} & \textbf{Weather} & \textbf{Traffic}\\
\midrule
\multirow{4}{*}{SAMformer} 
& 96   & 0.5 & 0.5 & 0.6 & 0.2  & 0.4 & 0.8 \\
& 192  & 0.6 & 0.8 & 0.9 & 0.9  & 0.4 & 0.1\\
& 336  & 0.9 & 0.6 & 0.9 & 0.8  & 0.6 & 0.5\\
& 720  & 0.9 & 0.8 & 0.9 & 0.9 & 0.5 & 0.7\\
\midrule
\multirow{4}{*}{TSMixer} 
& 96   & 1.0 & 0.9 & 1.0 & 1.0 &  0.5 & 0\\
& 192  & 0.7 & 0.1 & 0.6 & 1.0 &  0.4 & 0.9\\
& 336  & 0.7 & 0.0 & 0.7 & 1.0 &  0.6 & 0.6\\
& 720  & 0.3 & 0.4 & 0.5 & 1.0 &  0.3 & 0.9\\
\midrule
\multirow{4}{*}{FaCTR} 
& 96   & 0.85 & 0.8 & 0.6 & 0.75 &  0.7 & 0.05 \\
& 192  & 0.77 & 0.65 & 0.75 & 0.85 &   0.75 & 0.15\\
& 336  & 0.9 & 0.75 & 0.25 & 0.4 &   0.45 & 0.05\\
& 720  & 0.95 & 0.95 & 0.4 & 0.4 &   0.9 & 0.25\\
\bottomrule
\end{tabular}
\end{table}

\begin{table}[h]
\centering
\caption{Neighborhood size $\rho$ used in FaCTR across different datasets and prediction horizons for short-term forecasting datasets.}
\label{tab:rho_values_pems}
\begin{tabular}{llcccccccc}
\toprule
\textbf{Model} & \textbf{Horizon} & \textbf{PEMS03} & \textbf{PEMS04} & \textbf{PEMS07} & \textbf{PEMS08}\\
\midrule
\multirow{4}{*}{FaCTR} 
& 12   & 0.05 & 0.1 & 0.05 & 0.05\\
& 24  & 0.1 & 0.1 & 0.05 & 0.2\\
& 48  & 0.05 & 0.05 & 0.1 & 0.1 \\
& 96  & 0.2 & 0.05 & 0.1 & 0.15\\
\bottomrule
\end{tabular}
\end{table}

\subsubsection{Baseline details}
\label{appendix:modelperf}

In Table 2 (Section 4.2), we benchmark the performance of FaCTR against a diverse set of state-of-the-art baselines. For consistency, we fix the input (context) length to 512 and evaluate performance at multiple prediction horizons.

To ensure fairness and correctness in comparison, we recompute the performance of PatchTST, DLinear, and ModernTCN, as prior reported results were affected by known bugs in the original implementations\footnote{Not all samples were considered in the test dataloader \href{https://github.com/yuqinie98/PatchTST/blob/204c21efe0b39603ad6e2ca640ef5896646ab1a9/PatchTST_supervised/data_provider/data_factory.py\#L19}{GitHub: PatchTST data\_factory.py (line 19)}}. We adopt the results of SAMFormer and TSMixer directly from the official implementation reported ~\cite{ilbert2024}. Performance numbers for CrossGNN, Moment, and CSFormer are taken from 
~\cite{zhou2024moderntcn}, ~\cite{goswami2024moment} and ~\cite{zhang2025csformer} 
respectively. For each dataset, we use the standard splits for training, validation, and testing, as established in prior literature. All inputs are normalized using the mean and standard deviation computed from the training set, consistent with the preprocessing protocol introduced in ~\cite{wu2021autoformer}.

\subsection{Experimentation Results}
\subsubsection{Model Parameter Comparison}
\label{appendix:paramsapp}
Please refer to Table~\ref{tab:model_size} for a detailed view of parameters across datasets, horizons and models.\footnote{All numbers are calculated using official released repositories.}

\begin{table}[ht]
\centering
\scriptsize 
\caption{Model size comparison (parameter count) across datasets and forecast horizons. All models use a lookback window of 512. Values are reported in number of trainable parameters; \textcolor{blue}{blue} indicates the most compact model per row. FaCTR consistently exhibits the lowest parameter footprint across datasets and horizons.}
\label{tab:model_size}
\resizebox{\textwidth}{!}{%
\begin{tabular}{llrrrrrrrrr}
\toprule
\textbf{Dataset} & \textbf{Horizon} & \textbf{FaCTR} & \textbf{PatchTST} & \textbf{DLinear} & \textbf{TSMixer} & \textbf{CrossFormer} & \textbf{iTransformer} & \textbf{ModernTCN} & \textbf{Fedformer} & \textbf{AutoFormer} \\
\midrule
\multirow{4}{*}{ETTh2}
& 96  & \textcolor{blue}{71{,}296}   & 115{,}872   & 98{,}496   & 576{,}604   & 11{,}811{,}096 & 224{,}224   & 656{,}260    & 16{,}827{,}415 & 10{,}535{,}957 \\
& 192 & \textcolor{blue}{120{,}544} & 214{,}272   & 196{,}992  & 625{,}948   & 11{,}336{,}544 & 236{,}608   & 1{,}172{,}452 & 16{,}827{,}415 & 10{,}535{,}957 \\
& 336 & \textcolor{blue}{194{,}416} & 361{,}872   & 344{,}736  & 699{,}628   & 11{,}408{,}736 & 255{,}184   & 1{,}946{,}740 & 16{,}827{,}415 & 10{,}535{,}957\\
& 720 & 391{,}408                  & 755{,}472   & 738{,}720  & 896{,}620   & 11{,}601{,}248 & \textcolor{blue}{304{,}720} & 4{,}011{,}508 & 16{,}827{,}415 & 10{,}535{,}957 \\
\midrule
\multirow{4}{*}{Weather}
& 96  & \textcolor{blue}{71{,}744}   & 1{,}194{,}336 & 98{,}496   & 1{,}105{,}598 & 11{,}518{,}768 & 4{,}833{,}888 & 2{,}489{,}476 & 16{,}899{,}109 & 10{,}607{,}651 \\
& 192 & \textcolor{blue}{120{,}992} & 1{,}980{,}864 & 196{,}992  & 1{,}154{,}942 & 11{,}374{,}944 & 4{,}883{,}136 & 3{,}005{,}668 & 16{,}899{,}109 & 10{,}607{,}651 \\
& 336 & \textcolor{blue}{194{,}864} & 3{,}160{,}656 & 344{,}736  & 1{,}228{,}622 & 11{,}454{,}816 & 4{,}957{,}008 & 3{,}779{,}956 & 16{,}899{,}109 & 10{,}607{,}651 \\
& 720 & \textcolor{blue}{391{,}856} & 6{,}306{,}768 & 738{,}720  & 1{,}425{,}614 & 11{,}667{,}808 & 5{,}154{,}000 & 10{,}268{,}404 & 16{,}899{,}109 & 10{,}607{,}651 \\
\midrule
\multirow{4}{*}{Traffic}
& 96  & \textcolor{blue}{98{,}656}   & 921{,}187   & 98{,}496   & 3{,}042{,}412 & 3{,}589{,}744  & 6{,}411{,}872 & 822{,}756{,}868 & 21{,}205{,}870 & 14{,}914{,}412 \\
& 192 & \textcolor{blue}{147{,}904} & 1{,}437{,}379 & 196{,}992  & 3{,}091{,}756 & 2{,}411{,}424  & 6{,}461{,}120 & 824{,}231{,}524 & 21{,}205{,}870 & 14{,}914{,}412 \\
& 336 & \textcolor{blue}{221{,}776} & 2{,}211{,}667 & 344{,}736  & 3{,}165{,}436 & 3{,}219{,}616  & 6{,}534{,}992 & 826{,}443{,}508 & 21{,}205{,}870 & 14{,}914{,}412 \\
& 720 & \textcolor{blue}{418{,}768} & 4{,}276{,}435 & 738{,}720  & 3{,}362{,}428 & 3{,}681{,}440  & 6{,}534{,}992 & 832{,}342{,}132 & 21{,}205{,}870 & 14{,}914{,}412 \\
\midrule
\multirow{4}{*}{Electricity}
& 96  & \textcolor{blue}{81{,}344}   & 1{,}194{,}336 & 98{,}496   & 1{,}266{,}502 & 2{,}938{,}200  & 4{,}833{,}888 & 129{,}146{,}500 & 18{,}435{,}409 & 12{,}143{,}951 \\
& 192 & \textcolor{blue}{130{,}592} & 1{,}980{,}864 & 196{,}992  & 1{,}315{,}846 & 2{,}008{,}752  & 4{,}883{,}136 & 130{,}252{,}516 & 18{,}435{,}409 & 12{,}143{,}951 \\
& 336 & \textcolor{blue}{204{,}464} & 3{,}160{,}656 & 344{,}736  & 1{,}389{,}526 & 1{,}511{,}328  & 4{,}957{,}008 & 131{,}911{,}540 & 18{,}435{,}409 & 12{,}143{,}951 \\
& 720 & \textcolor{blue}{401{,}456} & 6{,}306{,}768 & 738{,}720  & 1{,}586{,}518 & 1{,}880{,}992  & 5{,}154{,}000 & 136{,}335{,}604 & 18{,}435{,}409 & 12{,}143{,}951 \\
\bottomrule
\end{tabular}
}
\end{table}











\subsubsection{Additional Long-Term Forecasting Performance Comparison}
\label{appendix:modelperformance}
To enable a more comprehensive comparison across a broader range of architectures, we additionally report long-term forecasting results from other models as presented in the literature. These supplementary baselines are included alongside our core comparisons to provide a fuller evaluation landscape. Please see Table \ref{tab:additionalperf} for details.
\begin{table}[h]
\centering
\caption{Long-Term Forecasting performance (MSE and MAE) on the datasets across models and horizons. Blue is best, red is second best}
\label{tab:additionalperf}
\resizebox{\textwidth}{!}{%
\begin{tabular}{llcccccccccccccccccccc}
\toprule
\textbf{Dataset} & \textbf{Horizon} & \multicolumn{2}{c}{\textbf{FaCTR}} & \multicolumn{2}{c}{\textbf{Crossformer}} & \multicolumn{2}{c}{\textbf{MTGNN}} & \multicolumn{2}{c}{\textbf{MTSMixer}} & \multicolumn{2}{c}{\textbf{TimesNet}} & \multicolumn{2}{c}{\textbf{Informer}} & \multicolumn{2}{c}{\textbf{Autoformer}} & \multicolumn{2}{c}{\textbf{FEDFormer}} & \multicolumn{2}{c}{\textbf{PyraFormer}} & \multicolumn{2}{c}{\textbf{LogTrans}} \\
\cmidrule(lr){3-4} \cmidrule(lr){5-6} \cmidrule(lr){7-8} \cmidrule(lr){9-10} \cmidrule(lr){11-12} \cmidrule(lr){13-14} \cmidrule(lr){15-16} \cmidrule(lr){17-18} \cmidrule(lr){19-20} \cmidrule(lr){21-22} 
& & MSE & MAE & MSE & MAE & MSE & MAE & MSE & MAE & MSE & MAE & MSE & MAE & MSE & MAE & MSE & MAE & MSE & MAE & MSE & MAE \\
\midrule
Weather & 96 & \textcolor{blue}{0.148} & \textcolor{blue}{0.200} & 0.174 & 0.214 & 0.230 & 0.329 & \textcolor{red}{0.156} & \textcolor{red}{0.206} & 0.172 & 0.220 & 0.354 & 0.405 & 0.249 & 0.329 & 0.238 & 0.314 & 0.896 & 0.556 & 0.458 & 0.490 \\
 & 192 & \textcolor{blue}{0.191} & \textcolor{blue}{0.240} & 0.221 & 0.254 & 0.263 & 0.322 & \textcolor{red}{0.199} & \textcolor{red}{0.248} & 0.219 & 0.261 & 0.419 & 0.434 & 0.325 & 0.370 & 0.275 & 0.329 & 0.622 & 0.624 & 0.658 & 0.589 \\
 & 336 & \textcolor{blue}{0.242} & \textcolor{blue}{0.280} & 0.278 & 0.296 & 0.354 & 0.396 & \textcolor{red}{0.249} & \textcolor{red}{0.291} & 0.280 & 0.306 & 0.583 & 0.543 & 0.351 & 0.391 & 0.339 & 0.377 & 0.739 & 0.753 & 0.797 & 0.652 \\
 & 720 & \textcolor{blue}{0.315} & \textcolor{blue}{0.332} & 0.358 & 0.347 & 0.409 & 0.371 & \textcolor{red}{0.336} & \textcolor{red}{0.343} & 0.365 & 0.359 & 0.916 & 0.705 & 0.415 & 0.426 & 0.389 & 0.409 & 1.004 & 0.934 & 0.869 & 0.675 \\
 \midrule
Traffic & 96 & \textcolor{blue}{0.379} & \textcolor{blue}{0.268} & \textcolor{red}{0.395} & \textcolor{red}{0.268} & 0.660 & 0.437 & 0.462 & 0.332 & 0.593 & 0.321 & 0.733 & 0.410 & 0.597 & 0.371 & 0.576 & 0.359 & 2.085 & 0.468 & 0.684 & 0.384 \\
 & 192 & \textcolor{blue}{0.395} & \textcolor{blue}{0.275} & \textcolor{red}{0.417} & \textcolor{red}{0.276} & 0.649 & 0.438 & 0.488 & 0.354 & 0.617 & 0.336 & 0.777 & 0.435 & 0.607 & 0.382 & 0.610 & 0.380 & 0.867 & 0.467 & 0.685 & 0.390 \\
 & 336 & \textcolor{blue}{0.404} & \textcolor{blue}{0.280} & \textcolor{red}{0.433} & \textcolor{red}{0.283} & 0.653 & 0.472 & 0.498 & 0.360 & 0.629 & 0.336 & 0.776 & 0.434 & 0.623 & 0.387 & 0.608 & 0.375 & 0.869 & 0.469 & 0.734 & 0.408 \\
 & 720 & \textcolor{blue}{0.437} & \textcolor{blue}{0.298} & \textcolor{red}{0.467} & \textcolor{red}{0.302} & 0.639 & 0.437 & 0.529 & 0.370 & 0.640 & 0.350 & 0.827 & 0.466 & 0.639 & 0.395 & 0.621 & 0.375 & 0.881 & 0.473 & 0.717 & 0.396 \\
 \midrule
Electricity & 96 & \textcolor{blue}{0.130} & \textcolor{blue}{0.226} & 0.148 & \textcolor{red}{0.240} & 0.217 & 0.318 & \textcolor{red}{0.141} & 0.243 & 0.168 & 0.272 & 0.304 & 0.393 & 0.196 & 0.313 & 0.186 & 0.302 & 0.386 & 0.449 & 0.258 & 0.357 \\
 & 192 & \textcolor{blue}{0.149} & \textcolor{blue}{0.243} & \textcolor{red}{0.162} & \textcolor{red}{0.253} & 0.238 & 0.352 & 0.163 & 0.261 & 0.184 & 0.289 & 0.327 & 0.417 & 0.211 & 0.324 & 0.197 & 0.311 & 0.386 & 0.443 & 0.266 & 0.368 \\
 & 336 & \textcolor{blue}{0.165} & \textcolor{blue}{0.260} & 0.178 & \textcolor{red}{0.269} & 0.260 & 0.348 & \textcolor{red}{0.176} & 0.277 & 0.198 & 0.300 & 0.333 & 0.422 & 0.214 & 0.327 & 0.213 & 0.328 & 0.378 & 0.443 & 0.280 & 0.380 \\
 & 720 & \textcolor{blue}{0.204} & \textcolor{blue}{0.293} & 0.225 & 0.317 & 0.290 & 0.369 & \textcolor{red}{0.212} & \textcolor{red}{0.308} & 0.220 & 0.320 & 0.351 & 0.427 & 0.236 & 0.342 & 0.233 & 0.344 & 0.376 & 0.445 & 0.283 & 0.376 \\
\midrule
ETTh1 & 96 & \textcolor{blue}{0.360} & \textcolor{blue}{0.390} & 0.386 & 0.405 & 0.515 & 0.517 & \textcolor{red}{0.372} & \textcolor{red}{0.395} & 0.384 & 0.402 & 0.941 & 0.769 & 0.435 & 0.446 & 0.376 & 0.399 & 0.664 & 0.612 & 0.878 & 0.740 \\
 & 192 & \textcolor{blue}{0.396} & \textcolor{blue}{0.412} & 0.441 & 0.436 & 0.553 & 0.522 & \textcolor{red}{0.416} & 0.426 & 0.557 & 0.436 & 1.007 & 0.786 & 0.456 & 0.457 & 0.423 & \textcolor{red}{0.416} & 0.790 & 0.681 & 1.037 & 0.824 \\
 & 336 & \textcolor{blue}{0.420} & \textcolor{blue}{0.429} & 0.487 & 0.458 & 0.612 & 0.577 & 0.455 & 0.449 & 0.491 & 0.469 & 1.038 & 0.784 & 0.486 & 0.487 & \textcolor{red}{0.444} & \textcolor{red}{0.443} & 0.891 & 0.738 & 1.238 & 0.932 \\
 & 720 & \textcolor{blue}{0.448} & \textcolor{blue}{0.460} & 0.503 & 0.491 & 0.609 & 0.597 & 0.475 & \textcolor{red}{0.472} & 0.521 & 0.500 & 1.144 & 0.857 & 0.515 & 0.517 & \textcolor{red}{0.469} & 0.490 & 0.963 & 0.782 & 1.135 & 0.852 \\
\midrule
ETTh2 & 96 & \textcolor{blue}{0.274} & \textcolor{blue}{0.338} & \textcolor{red}{0.297} & \textcolor{red}{0.349} & 0.354 & 0.454 & 0.307 & 0.354 & 0.340 & 0.374 & 1.549 & 0.952 & 0.332 & 0.368 & 0.332 & 0.374 & 0.645 & 0.597 & 2.116 & 1.197 \\
 & 192 & \textcolor{blue}{0.337} & \textcolor{blue}{0.379} & 0.380 & 0.400 & 0.457 & 0.464 & \textcolor{red}{0.374} & \textcolor{red}{0.399} & 0.402 & 0.414 & 3.792 & 1.542 & 0.426 & 0.434 & 0.407 & 0.446 & 0.788 & 0.683 & 4.315 & 1.635 \\
 & 336 & \textcolor{blue}{0.360} & \textcolor{blue}{0.410} & 0.428 & \textcolor{red}{0.432} & 0.515 & 0.540 & \textcolor{red}{0.398} & 0.432 & 0.452 & 0.452 & 4.215 & 1.642 & 0.477 & 0.479 & 0.400 & 0.447 & 0.907 & 0.747 & 1.124 & 1.604 \\
 & 720 & \textcolor{blue}{0.398} & \textcolor{blue}{0.434} & 0.427 & \textcolor{red}{0.445} & 0.532 & 0.576 & 0.463 & 0.465 & 0.462 & 0.468 & 3.656 & 1.619 & 0.453 & 0.490 & \textcolor{red}{0.412} & 0.469 & 0.963 & 0.783 & 3.188 & 1.540 \\
 \midrule
ETTm1 & 96 & \textcolor{blue}{0.284} & \textcolor{blue}{0.338} & 0.334 & 0.368 & 0.379 & 0.446 & \textcolor{red}{0.314} & \textcolor{red}{0.358} & 0.338 & 0.375 & 0.626 & 0.560 & 0.510 & 0.492 & 0.326 & 0.390 & 0.543 & 0.510 & 0.600 & 0.546 \\
 & 192 & \textcolor{blue}{0.327} & \textcolor{blue}{0.362} & 0.377 & 0.391 & 0.470 & 0.428 & \textcolor{red}{0.354} & \textcolor{red}{0.386} & 0.371 & 0.387 & 0.725 & 0.619 & 0.514 & 0.495 & 0.365 & 0.415 & 0.557 & 0.537 & 0.837 & 0.700 \\
 & 336 & \textcolor{blue}{0.360} & \textcolor{blue}{0.388} & 0.426 & 0.420 & 0.473 & 0.430 & \textcolor{red}{0.384} & \textcolor{red}{0.405} & 0.410 & 0.411 & 1.005 & 0.741 & 0.510 & 0.492 & 0.392 & 0.425 & 0.754 & 0.655 & 1.124 & 0.832 \\
 & 720 & \textcolor{blue}{0.414} & \textcolor{blue}{0.416} & 0.491 & 0.459 & 0.553 & 0.479 & \textcolor{red}{0.427} & \textcolor{red}{0.432} & 0.478 & 0.450 & 1.133 & 0.845 & 0.527 & 0.493 & 0.446 & 0.458 & 0.908 & 0.724 & 1.153 & 0.820 \\
\midrule
ETTm2 & 96 & \textcolor{blue}{0.163} & \textcolor{blue}{0.252} & 0.180 & 0.264 & 0.203 & 0.299 & \textcolor{red}{0.177} & \textcolor{red}{0.259} & 0.187 & 0.267 & 0.355 & 0.462 & 0.205 & 0.293 & 0.180 & 0.271 & 0.435 & 0.507 & 0.768 & 0.642 \\
 & 192 & \textcolor{blue}{0.217} & \textcolor{blue}{0.290} & 0.250 & 0.309 & 0.265 & 0.328 & \textcolor{red}{0.241} & \textcolor{red}{0.303} & 0.249 & 0.309 & 0.595 & 0.586 & 0.278 & 0.336 & 0.252 & 0.318 & 0.730 & 0.673 & 0.989 & 0.757 \\
 & 336 & \textcolor{blue}{0.266} & \textcolor{blue}{0.323} & 0.311 & 0.348 & 0.365 & 0.374 & \textcolor{red}{0.297} & \textcolor{red}{0.338} & 0.321 & 0.351 & 1.270 & 0.871 & 0.343 & 0.379 & 0.324 & 0.364 & 1.201 & 0.845 & 1.334 & 0.872 \\
 & 720 & \textcolor{blue}{0.346} & \textcolor{blue}{0.376} & 0.412 & 0.407 & 0.461 & 0.459 & \textcolor{red}{0.396} & \textcolor{red}{0.398} & 0.497 & 0.403 & 3.001 & 1.267 & 0.414 & 0.419 & 0.410 & 0.420 & 3.625 & 1.451 & 3.048 & 1.328 \\
\bottomrule
\end{tabular}
}
\end{table}

\subsubsection{Efficiency Comparison}
\label{appendix:efficiency}
\begin{table}[ht]
\centering
\caption{Training time and peak memory usage (per epoch) for different models across datasets. Context = 512, Prediction length = 96, Batch size = 32.}
\label{tab:memeff}
\resizebox{\textwidth}{!}{%
\begin{tabular}{llcc}
\toprule
\textbf{Dataset} & \textbf{Model} & \textbf{Training Time (per epoch)} & \textbf{Peak Memory Allocation (per epoch)} \\
\midrule
\multirow{5}{*}{ETTh2}
  & FaCTR      & 4s   & 45.61 MB   \\
  & PatchTST   & 4s   & 181.89 MB  \\
  & ModernTCN  & 2s   & 144.58 MB  \\
  & DLinear    & 2s   & 21.74 MB   \\
  & SAMFormer  & 2s   & 26.59 MB   \\
\midrule
\multirow{5}{*}{Weather}
  & FaCTR      & 12s  & 87.39 MB   \\
  & PatchTST   & 41s  & 2023.41 MB \\
  & ModernTCN  & 43s  & 1227.61 MB \\
  & DLinear    & 6s   & 27.77 MB   \\
  & SAMFormer  & 8s   & 32.87 MB   \\
\midrule
\multirow{5}{*}{Electricity}
  & FaCTR      & 30s  & 800.22 MB \\
  & PatchTST   & 272s & 30116.95 MB \\
  & ModernTCN  & 446s & 19555.58 MB \\
  & DLinear    & 15s  & 156.24 MB  \\
  & SAMFormer  & 8s   & 205.84 MB  \\
\bottomrule
\end{tabular}
}
\end{table}

To assess the computational efficiency of our model, we compare the training time per epoch and peak memory usage across datasets and models, as shown in Table~\ref{appendix:efficiency}. For consistency, we fix the context length to 512, prediction length to 96, and batch size to 32, ensuring all models are evaluated on the same task and under identical conditions. We also record the peak memory allocated on the GPU over all batches during one epoch. Note that we perform this analysis for the top 5 performing models.

Our results demonstrate that FaCTR is significantly more lightweight than many state-of-the-art models such as PatchTST and ModernTCN, both in terms of memory consumption and training speed. This supports our central hypothesis: effective time series forecasting does not require excessively heavy, over-parameterized architectures.

While FaCTR may be marginally more memory-intensive than the simplest models like DLinear, it strikes a compelling balance between efficiency and performance. Further, FaCTR converges substantially faster than other models in practice. For example, ETT datasets typically converge in under 50 epochs, and larger datasets such as Traffic, Weather, and Electricity reach convergence in 80–100 epochs with early stopping. In contrast, heavier models often require longer training durations and greater GPU memory.

These findings highlight FaCTR's capability to offer strong predictive performance with competitive training efficiency, making it a practical and scalable solution for real-world time series forecasting applications.

\subsubsection{Parameter Efficiency Comparison with Crossformer}
\label{appendix:efficiencymul}

To quantify the parameter efficiency of FaCTR, we compute the Crossformer-to-FaCTR parameter ratio across 16 dataset–horizon combinations (ETTh2, Weather, Traffic, Electricity; horizons 96, 192, 336, 720). Ratios range from $4.7\times$ (Electricity, 720) to $165.7\times$ (ETTh2, 96). The average multiplier across all cases is \textbf{53.8$\times$}, confirming that FaCTR achieves competitive performance with significantly fewer parameters.

\subsubsection{Forecast Prediction Examples}
\label{appendix:forecasts}

To qualitatively evaluate the performance of our model, we visualize forecasted values against ground truth across different datasets and prediction horizons. The following plots illustrate how FaCTR effectively captures trends and seasonal patterns across multiple channels. The forecasts closely align with the actual values, demonstrating the model’s ability to track both long-term dependencies and short-term fluctuations. This level of fidelity is consistent across a wide range of prediction horizons, further validating the model’s robustness and generalization capacity. \\

\textbf{ETTh datasets}
\begin{figure}[H]
    \centering
    \includegraphics[width=0.85\linewidth]{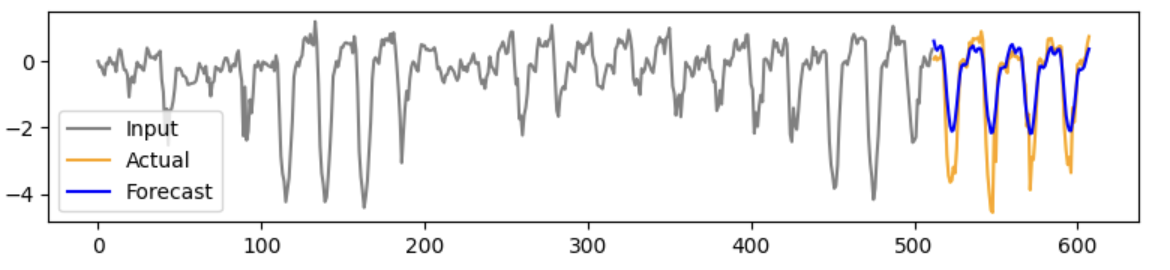}
    \caption{Forecast vs. actual plot for channel 0 in ETTh1 for prediction horizon 96}
    \label{fig:forecast_example1}
\end{figure}
\vspace{-5em}
\begin{figure}[H]
    \centering
    \includegraphics[width=0.85\linewidth]{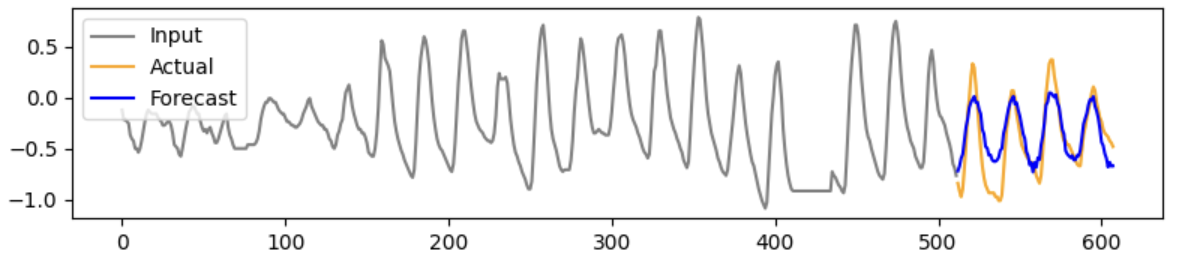}
    \caption{Forecast vs. actual plot for channel 6 in ETTh2 for prediction horizon 96}
    \label{fig:forecast_example2}
\end{figure}
\vspace{-2em}
\begin{figure}[H]
    \centering
    \includegraphics[width=0.85\linewidth]{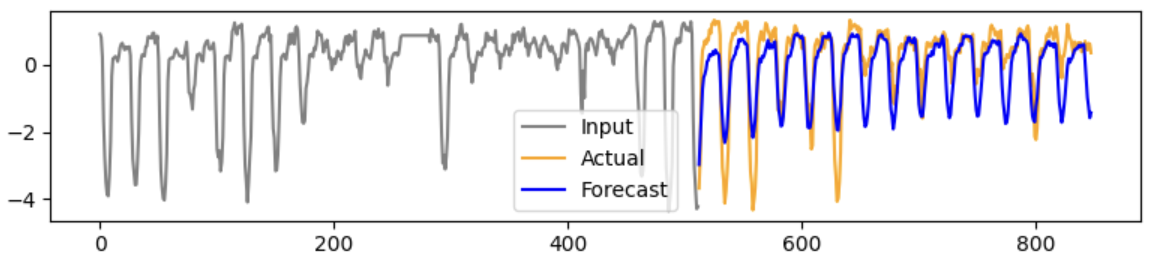}
    \caption{Forecast vs. actual plot for channel 4 in ETTh1 for prediction horizon 336}
    \label{fig:forecast_example3}
\end{figure}
\textbf{Weather}
\begin{figure}[H]
    \centering
    \includegraphics[width=0.85\linewidth]{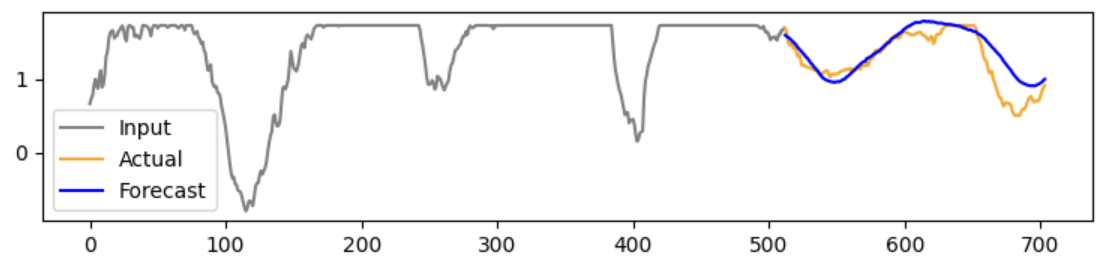}
    \caption{Forecast vs. actual plot for channel 16 in Weather for prediction horizon 192}
    \label{fig:forecast_example5}
\end{figure}
\vspace{-2em}
\begin{figure}[H]
    \centering
    \includegraphics[width=0.85\linewidth]{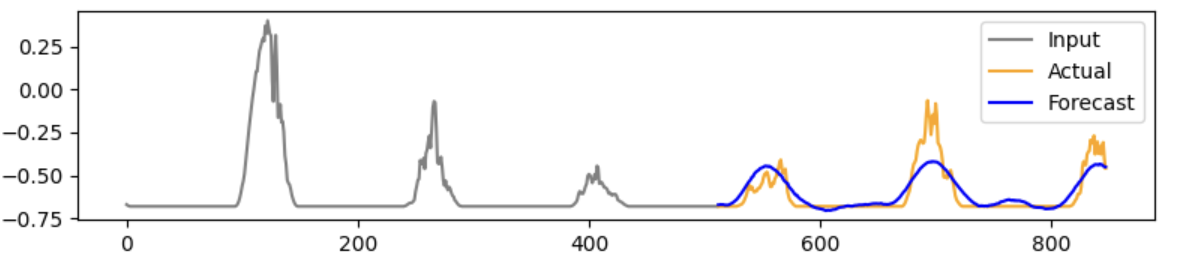}
    \caption{Forecast vs. actual plot for channel 2 in Weather for prediction horizon 336}
    \label{fig:forecast_example4}
\end{figure}

\textbf{Electricity}

\begin{figure}[H]
    \centering
    \includegraphics[width=0.85\linewidth]{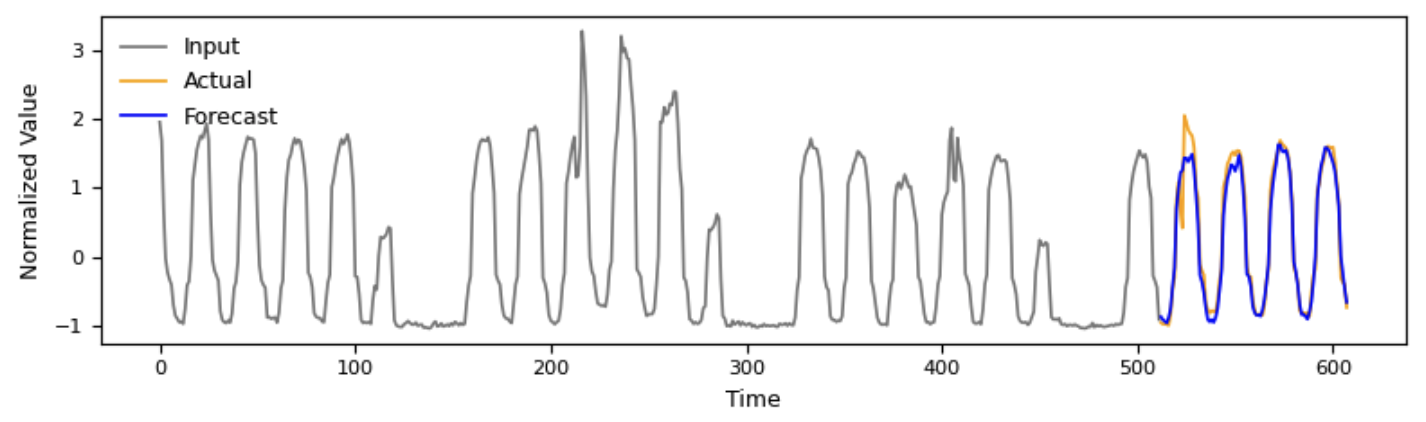}
    \caption{Forecast vs. actual plot for channel 28 in Electricity for prediction horizon 96}
    \label{fig:forecast_example9}
\end{figure}

\begin{figure}[H]
    \centering
    \includegraphics[width=0.85\linewidth]{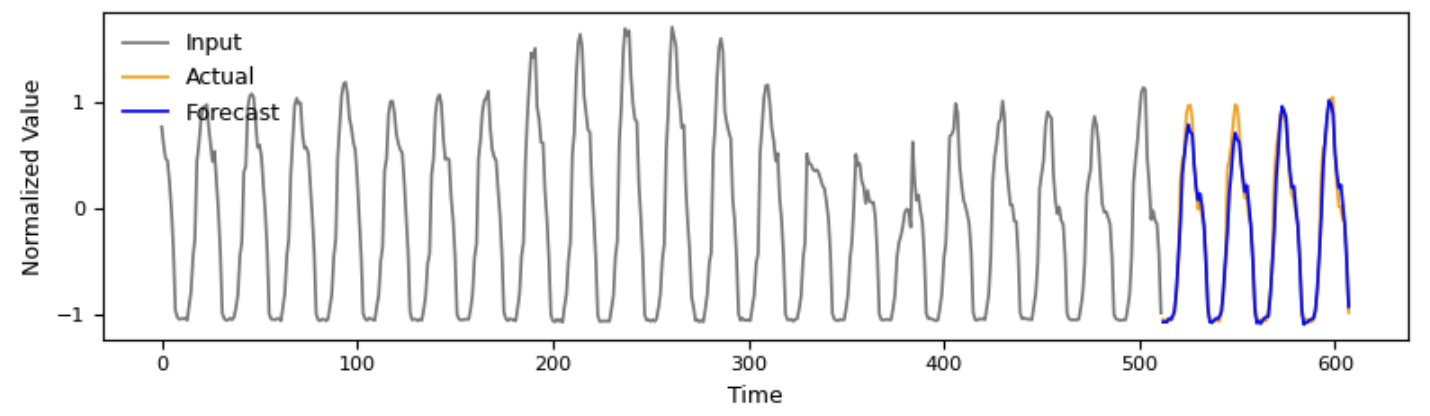}
    \caption{Forecast vs. actual plot for channel 310 in Electricity for prediction horizon 96}
    \label{fig:forecast_example10}
\end{figure}

\vspace{-0.5em}
\textbf{Traffic}
\vspace{-0.5em}
\begin{figure}[H]
    \centering
    \includegraphics[width=0.85\linewidth]{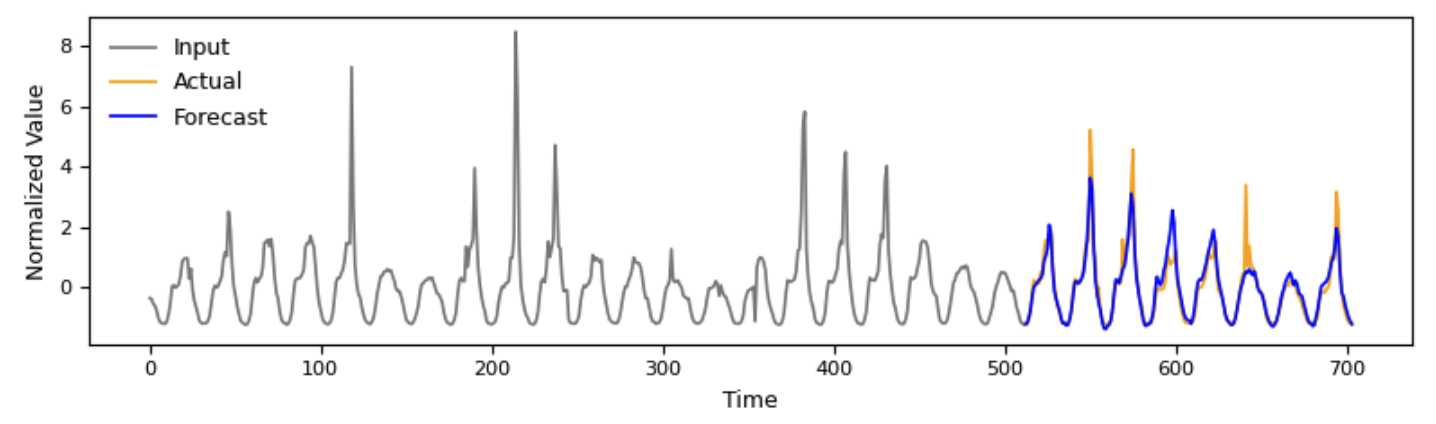}
    \caption{Forecast vs. actual plot for channel 3 in Traffic for prediction horizon 192}
    \label{fig:forecast_example11}
\end{figure}
\vspace{-4.5em}
\begin{figure}[H]
    \centering
    \includegraphics[width=0.85\linewidth]{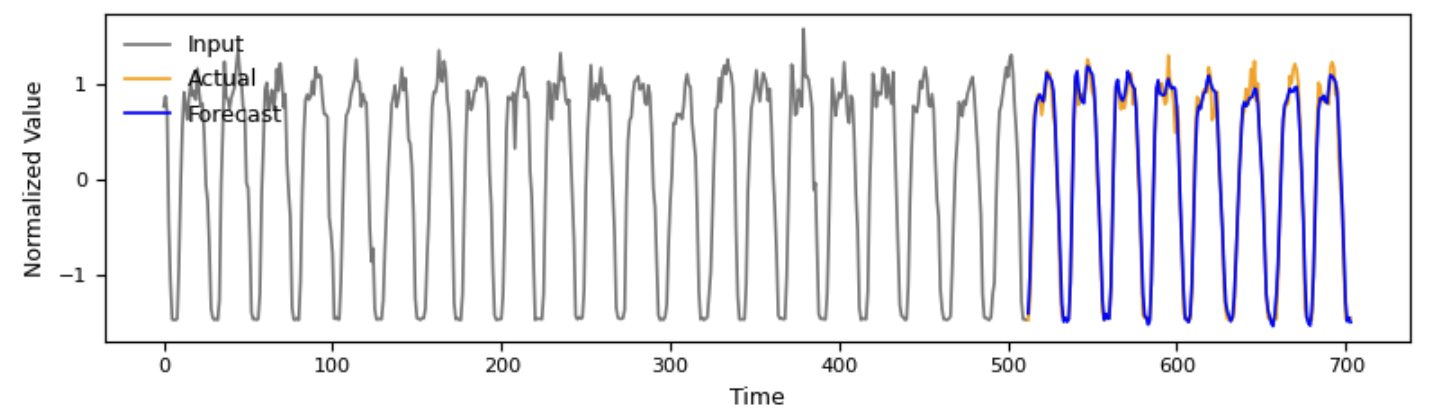}
    \caption{Forecast vs. actual plot for channel 86 in Traffic for prediction horizon 192}
    \label{fig:forecast_example12}
\end{figure}

\subsubsection{Diagnosing underperformance on the Traffic Dataset}
Despite competitive overall performance, our model exhibits some underperformance on the Traffic dataset, when compared to PatchTST. To identify performance bottlenecks, we computed the Mean Squared Error (MSE) and Mean Absolute Error (MAE) for each channel independently across all samples and prediction horizons.

\paragraph{Channel-Wise Error Analysis.}
Table~\ref{tab:channel_errors} reveals that a small subset of channels contributes disproportionately to the overall error. In particular, channel 824 exhibits a markedly high MSE of 44.81—nearly an order of magnitude larger than the next worst channel (520 with MSE 5.30)—indicating a localized failure mode in the model's predictions.

\begin{table}[H]
\centering
\caption{Top 10 channels with the highest forecast error (ranked by MSE).}
\label{tab:channel_errors}
\begin{tabular}{ccc}
\toprule
\textbf{Channel} & \textbf{MSE} & \textbf{MAE} \\
\midrule
824 & 44.813950 & 3.686094 \\
520 & 5.300683  & 0.973157 \\
693 & 4.740004  & 0.787192 \\
699 & 3.553118  & 0.878122 \\
604 & 3.433676  & 0.978763 \\
695 & 2.991475  & 0.781612 \\
460 & 2.582403  & 0.730914 \\
632 & 2.340947  & 0.736961 \\
525 & 1.813177  & 0.633519 \\
709 & 1.687670  & 0.646433 \\
\bottomrule
\end{tabular}
\end{table}

\paragraph{Forecast Errors.}
We analyze the predicted vs. actual trajectories for random samples for the worst-performing channels to better understand the sources of error. While the model captures broad seasonal patterns, it consistently fails to predict sharp peaks and tends to over-smooth high-frequency transitions. This suggests difficulty in modeling abrupt dynamics, which may arise from: (1) low signal variance during training, (2) sparse or noisy inputs, or (3) an inductive bias toward smooth temporal evolution.

\begin{figure}[H]
    \centering
    \begin{subfigure}[b]{0.45\textwidth}
        \centering
        \includegraphics[width=0.8\textwidth]{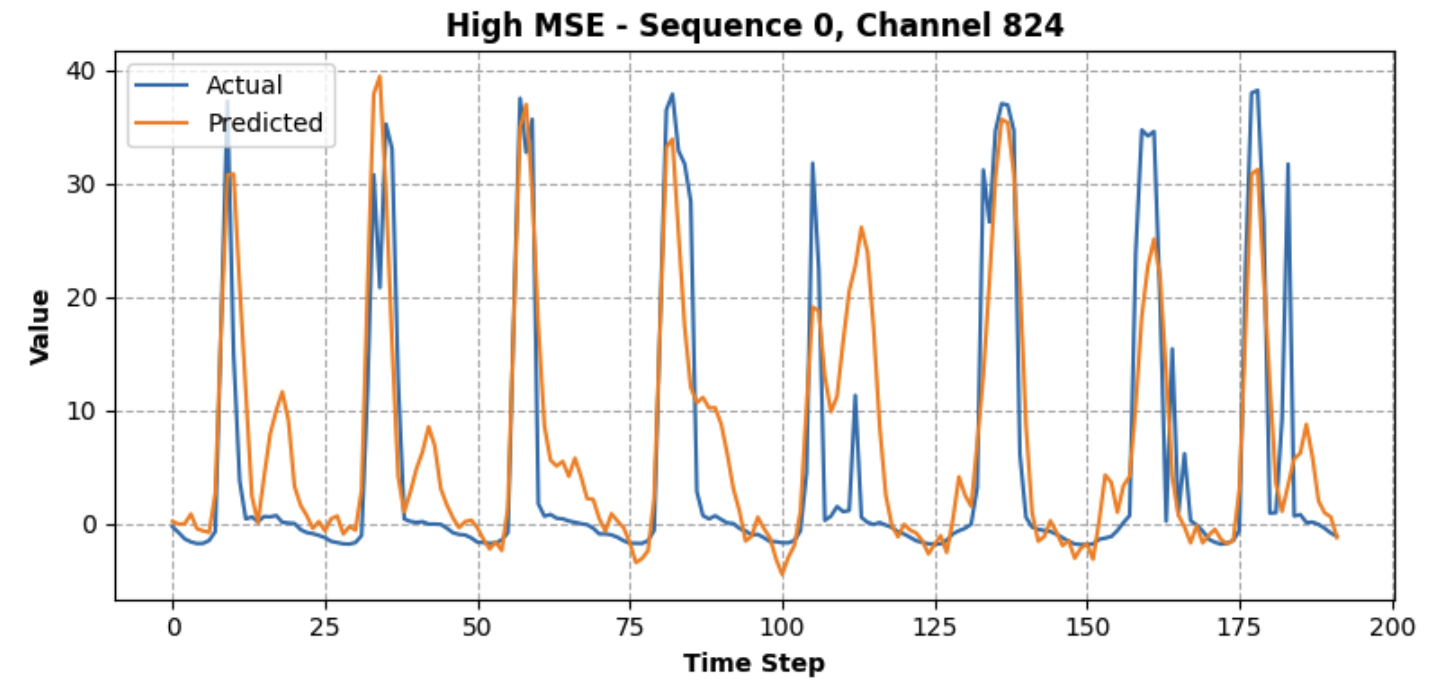}
        \caption{Predictions vs Actuals}
    \end{subfigure}
    
    \vspace{1em}  

    \begin{subfigure}[b]{0.8\textwidth}
        \centering
        \includegraphics[width=\textwidth]{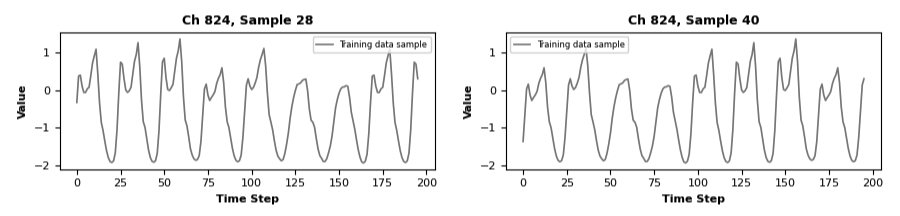}
        \caption{Training data samples}
    \end{subfigure}

    \caption{Examples from channels with distinct training-time patterns influencing forecast behavior. The model learns double peaks for channel 824.}
    \label{fig:pattern_bias_examples1}
\end{figure}

The model’s behavior appears tightly coupled to patterns prevalent in the training data. For instance, as can be seen in Figure \ref{fig:pattern_bias_examples1}, channel 824 frequently exhibits closely spaced peaks in its training sequences, leading the model to internalize a bias toward dual-peak structures. Consequently, it confidently outputs such motifs even when the test sequences deviate—highlighting a case of overgeneralization from repetitive patterns.

\begin{figure}[H]
    \centering
    \begin{subfigure}[b]{0.45\textwidth}
        \centering
        \includegraphics[width=0.8\textwidth]{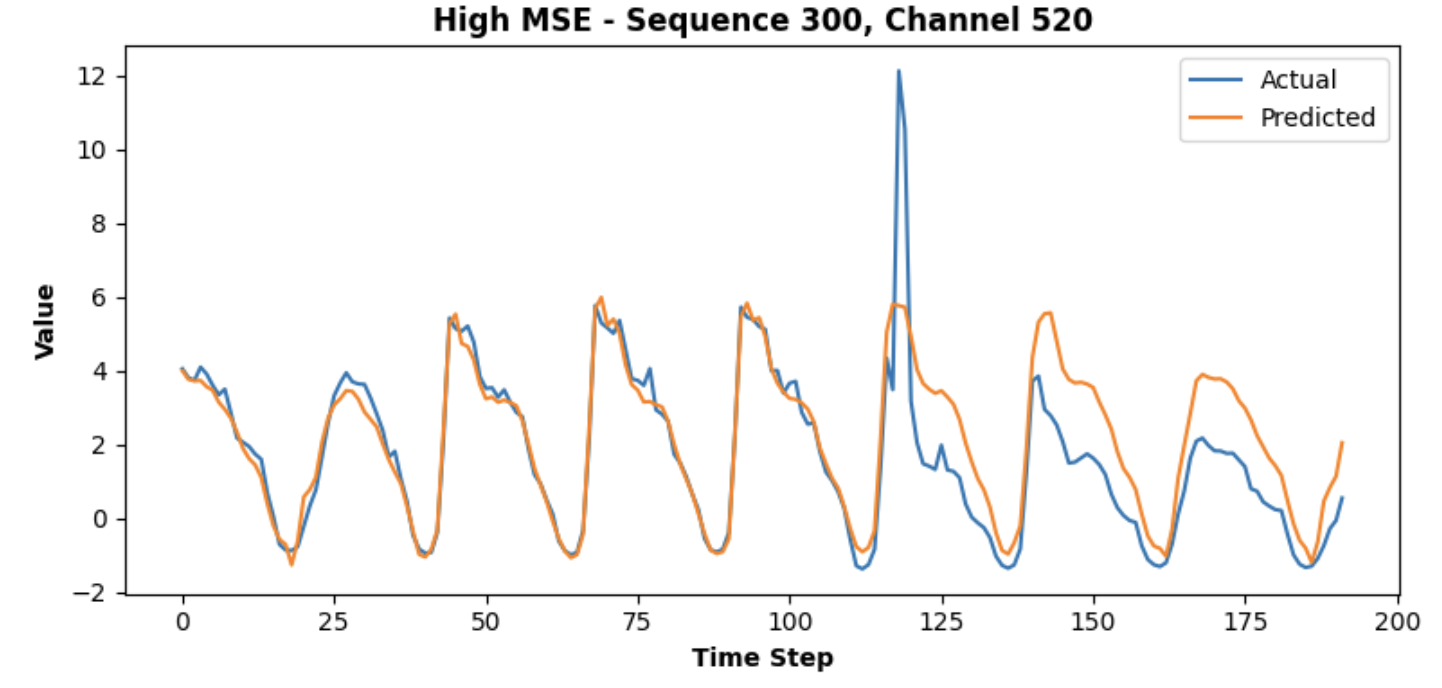}
        \caption{Predictions vs Actuals for channel 520}
    \end{subfigure}
    
    \vspace{1em}  

    \begin{subfigure}[b]{0.8\textwidth}
        \centering
        \includegraphics[width=\textwidth]{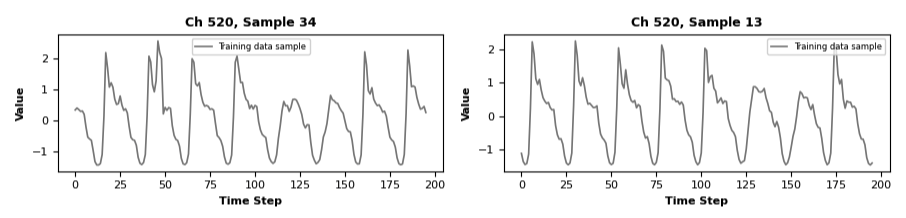}
        \caption{Training data samples}
    \end{subfigure}

    \caption{Examples from channels with different patterns from training-time. The model misses sudden peaks in channel 520.}
    \label{fig:pattern_bias_examples2}
\end{figure}

\vspace{-2em}
In contrast as seen in Figure \ref{fig:pattern_bias_examples2}, channel 520 is characterized by smooth seasonal cycles during training. The model becomes strongly conditioned on this regularity and fails to adapt when the test data introduces sudden spikes, continuing instead to predict the expected periodic behavior.

\begin{figure}[H]
    \centering
    \begin{subfigure}[b]{0.45\textwidth}
        \centering
        \includegraphics[width=0.8\textwidth]{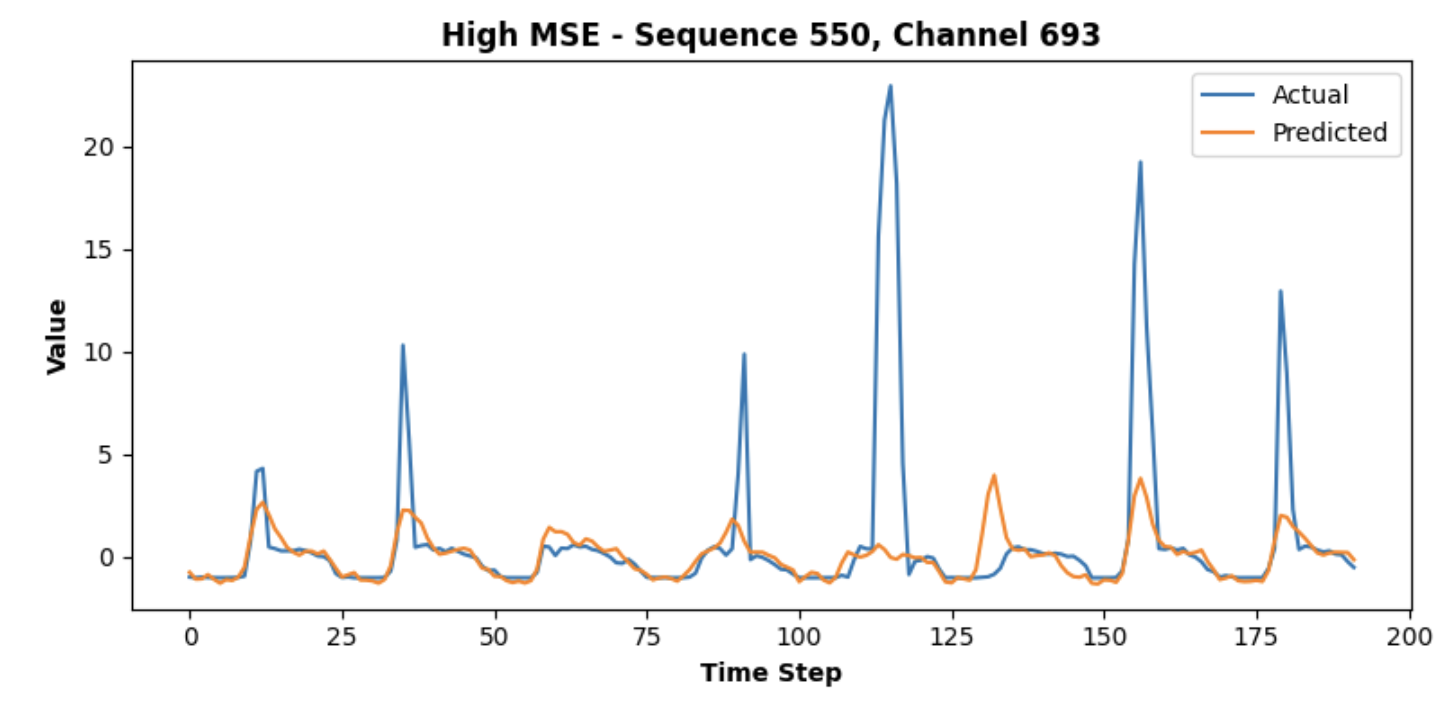}
        \caption{Predictions vs Actuals}
    \end{subfigure}
    
    \vspace{1em}  

    \begin{subfigure}[b]{1.0\textwidth}
        \centering
        \includegraphics[width=0.8\textwidth]{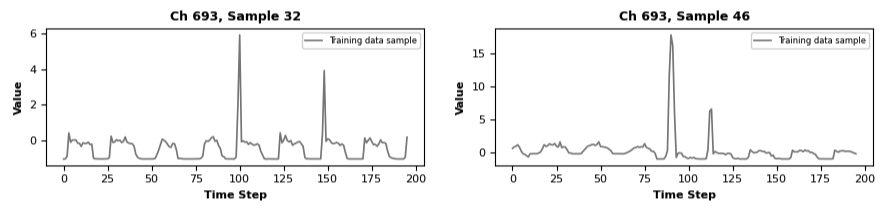}
        \caption{Training data samples}
    \end{subfigure}

    \caption{Examples from channels with distinct training-time patterns influencing forecast behavior. The model fails to predict peaks in channel 693 since it treats them as outliers.}
    \label{fig:pattern_bias_examples3}
\end{figure}

A similar issue arises in channel 693 as seen in Figure \ref{fig:pattern_bias_examples3}, where abrupt peaks occur during testing but are absent in the training data. The model, having never encountered such dynamics, defaults to conservative forecasts that underpredict or entirely miss these sharp transitions.

These observations underscore the importance of training on data that captures greater variability, including rare or extreme events. As part of our ongoing work, we plan to integrate these insights into architectural improvements aimed at enhancing robustness and reducing the long tail of poor channel-level performance.

\section{Representation Learning Experiment Details}
\label{appendix:replearn}

We adopt a self-supervised representation learning strategy for FaCTR by leveraging a masked patch reconstruction task on the ETTh1 dataset, thereby forcing the network to infer latent structure and temporal relationships. As mentioned in section 4.4, we mask 45\% of the input patches, and pre-train the model to reconstruct them. We use a patch size of 32 with a stride of 32, resulting in non-overlapping patches over a fixed context window of 512, consistent with the architectural setup used in our benchmarking experiments.

Furthermore, we perform linear probing by training a lightweight forecasting head for 10 epochs on top of the frozen backbone learned during pretraining. Additionally, in the end-to-end fine-tuning setup, we adopt a two-stage process: we first conduct linear probing for 10 epochs to warm-start the forecasting head, followed by 20 epochs of full fine-tuning where the entire model, including the encoder, is updated.

\subsection{Why Linear Probing and Fine-tuning demonstrate competant results}
ETTh1 and ETTh2 are related yet distinct multivariate time series datasets from the same family (Electricity Transformer Temperature), each collected from different transformers with potentially varying load conditions and patterns. Despite these differences, both datasets share similar covariates, and underlying physical processes governed by transformer dynamics.

Pretraining on ETTh1 allows FaCTR to learn spatiotemporal dependencies, periodicities, and variable interactions that are transferable to ETTh2. This transfer is particularly meaningful for two reasons:

\textbf{Inductive Bias via Patching and Masking:}
By operating over non-overlapping patches and applying high masking ratios (45\%), the model is compelled to learn global contextual information to reconstruct missing portions. This promotes robustness and generalization to unseen patterns in ETTh2.

\textbf{Shared Dynamics and Cross-Correlation Structures:}
Both datasets exhibit similar inter-channel correlations and temporal behaviors, even if the specific patterns differ. The masked reconstruction task effectively encourages the model to encode these correlations in a way that benefits supervised downstream tasks.

These factors together enable the pretrained FaCTR model to extract high-quality, general-purpose representations during self-supervised learning on ETTh1. As a result, even without updating the backbone, linear probing yields competitive performance on ETTh2 by leveraging features that capture essential temporal and cross-channel structures. This highlights the robustness and transferability of the learned representations, demonstrating the potential of FaCTR's lightweight architecture.

\subsection{Caveats of transfer learning in FaCTR}

While FaCTR benefits from self-supervised pretraining through masked patch reconstruction, applying transfer learning across domains presents a key architectural constraint. Specifically, FaCTR incorporates spatial embeddings that are directly tied to the number of input channels (e.g., sensors, entities, or time series variables). This introduces a limitation when transferring across datasets with different numbers or identities of input channels, which is common in real-world scenarios.

In typical transfer learning workflows, one may pretrain a model on a large source dataset and fine-tune it on a target dataset from a different domain. However, in FaCTR, if the number of channels (and thus spatial entities) changes between domains, the learned spatial embeddings from the source domain become incompatible with the target domain.

To address this, a practical strategy is to freeze the encoder (which captures temporal and structural representations) and retrain both the task head and the spatial embeddings on the target domain. This would allow the model to adapt to the new set of spatial entities while still leveraging the pretrained temporal representation backbone. 

\section{Case Study: Retail Demand Forecasting with Synthetic Multivariate Time Series}
\label{appendix:casestudy}

\subsection{Motivation}
In real-world retail forecasting, the demand for a product is rarely determined by its own historical sales alone. Instead, it is shaped by a complex interplay of seasonal trends, promotional effects, inter-product dependencies, market dynamics, and external disruptions. Accurately capturing these factors is critical for producing reliable and robust forecasts. However, evaluating a model’s ability to reason about such structure is challenging, especially in the absence of clean, labeled datasets with known causal relationships.

To address this gap, we construct a synthetic dataset that mirrors key behavioral patterns observed in retail demand data. Each channel in the dataset represents a distinct, interpretable demand signal — such as periodic seasonality, random noise, or cross-channel dependencies — allowing us to systematically evaluate how well a forecasting model can capture temporal patterns, infer inter-channel relationships, and generalize under structured dependencies.

\subsection{Scenario}

We simulate daily demand signals for 8 product categories (or retail stores/regions) in a synthetic retail chain over multiple years. Each channel $C_{i}$ is carefully designed to mimic a distinct and interpretable demand behavior commonly encountered in real-world forecasting tasks — including seasonality, trend, promotions, noise, and inter-product dependencies.

This design allows us to explicitly test whether models can capture not only temporal patterns but also cross-channel causal structures such as promotion spillovers, demand cannibalization, and lagged responses.

\begin{table}[h]
\centering
\caption{Channel design and simulated demand patterns. Each channel corresponds to a distinct signal type inspired by real-world retail dynamics.}
\label{tab:channel_breakdown}
\renewcommand{\arraystretch}{1.3}
\begin{tabularx}{\textwidth}{l l X}
\toprule
\textbf{Channel} & \textbf{Simulated Signal} & \textbf{Real-World Interpretation} \\
\midrule
C1 & Low-frequency sine wave & Stable weekly seasonality, e.g., demand for milk or eggs with consistent consumption patterns. \\
C2 & Duplicate of C1 & Perfectly correlated with C1. Represents a private-label alternative to C1 (same store, different brand). Used to test redundancy modeling. \\
C3 & High-frequency sine wave & Rapidly fluctuating demand, e.g., snack items with impulse-driven purchases. Evaluates responsiveness to high-frequency signals. \\
C4 & Seasonality + upward trend & Regular weekly cycles with a long-term growth trend, e.g., increasing sales of protein shakes amid rising health awareness. \\
C5 & Random Gaussian noise & Irregular, erratic behavior. Example: clearance or discontinued items with unpredictable demand. Tests robustness to noise. \\
C6 & Promotion signal (leading indicator) & External marketing or advertising campaign for a product. Models an upstream promotional driver that influences demand in other channels. \\
C7 & Demand increases after C6 & Product whose sales rise in response to C6's promotion. Example: a sports drink promoted via ads (C6) with demand peaking shortly after. Tests causal lag learning. \\
C8 & Demand suppression due to C7 & Competing product whose demand falls when C7 is promoted. Example: another energy drink cannibalized by C7's campaign. Captures competitive effects. \\
\bottomrule
\end{tabularx}
\end{table}

To visualize the nature of the input data received by the model, we show a representative multivariate sample from the synthetic dataset in Figure 2. This structure allows us to evaluate the model’s ability to leverage temporal context, cross-channel cues, and noise filtering in generating accurate forecasts.

\begin{figure}[h]
\centering
\includegraphics[width=1.0\linewidth]{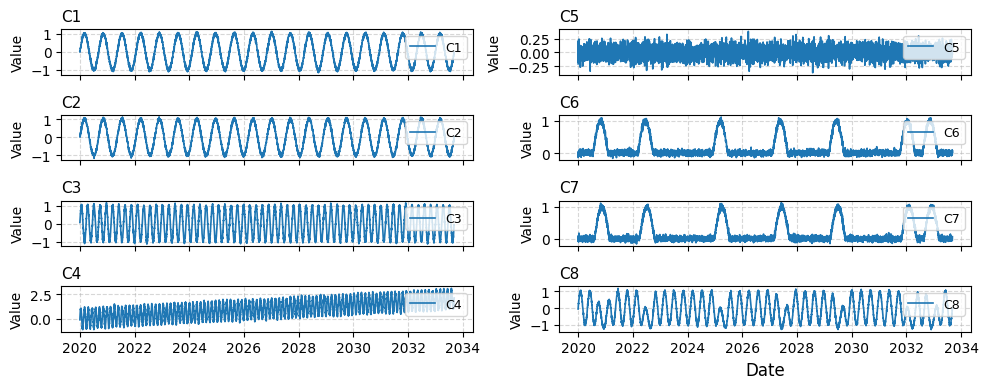}
\caption{Simulated time series patterns for each channel. Each channel exhibits a distinct signal structure (e.g., seasonality, noise, lag, or trend) designed to test specific modeling capabilities.}
\label{fig:channel_signals}
\end{figure}

\subsection{Model Forecasting Results}

To evaluate the forecasting capabilities of our model, we visualize its predictions across all channels. 

\subsubsection{Qualitative Observations Across Samples}
\begin{figure}[h]
\centering
\includegraphics[width=1.0\linewidth]{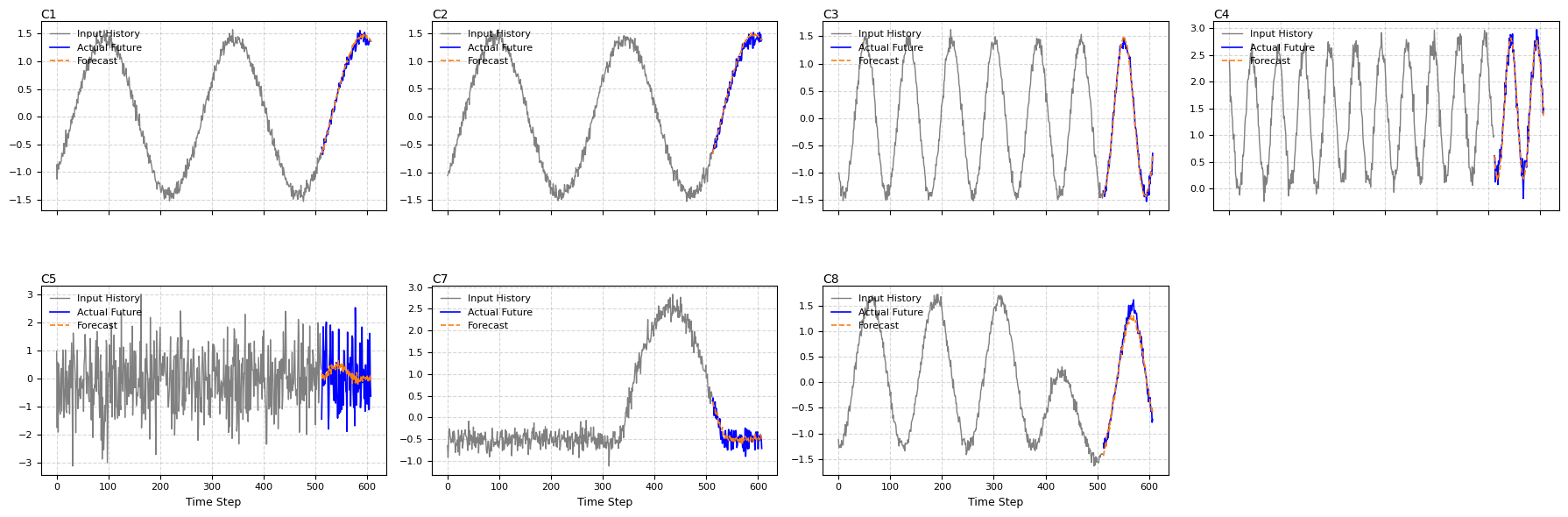}
\caption{Actuals vs Forecasts across channels}
\label{fig:forecast_all}
\end{figure}

The model demonstrates a strong ability to capture a range of temporal and cross-channel behaviors across the eight simulated demand signals. For C1 and C2, which follow stable sinusoidal patterns, the model generalizes across phases with high fidelity, reflecting robust seasonal baseline tracking. C3’s high-frequency oscillations are forecasted with precision, showcasing the model’s ability to learn local temporal dynamics and maintain phase and amplitude alignment. In C4, which combines an upward trend with weekly seasonality, the model accurately disentangles and extrapolates both components—highlighting its capability to model additive signals, as commonly seen in retail data. For C5, the model appropriately suppresses its response to unstructured noise, avoiding overfitting and producing smoothed forecasts—indicating robustness to irrelevant variance. More notably, in the C6–C7 interaction, the model captures a causal lag pattern, where C6 acts as a leading promotion signal and C7 follows with a delayed response, including the timing and shape of the response curve. Finally, C8 reflects a case of promotion-induced inverse cannibalization, where its demand rises back to normal in response to C6’s fall. The model successfully learns this inverse relationship, capturing the recovery in C8, demonstrating a nuanced understanding of cross-channel dependencies and competitive effects.

\end{document}